\documentclass{article}

 \usepackage[preprint]{neurips_2026}

\usepackage[utf8]{inputenc} 
\usepackage[T1]{fontenc}    
\usepackage{hyperref}       
\usepackage{url}            
\usepackage{booktabs}       
\usepackage{amsfonts}       
\usepackage{nicefrac}       
\usepackage{microtype}      
\usepackage{xcolor}         
\usepackage{xspace}
\usepackage{subcaption}
\usepackage{graphicx}
\usepackage{diagbox}
\usepackage{multirow} 
\usepackage{graphicx}
\usepackage{subcaption}
\usepackage{booktabs} 
\usepackage{diagbox}
\usepackage{multirow} 
\usepackage[table]{xcolor}
\usepackage{soul}
\usepackage{makecell}
\usepackage{amsmath}
\usepackage{fontawesome5}

\RequirePackage{algorithm}
\RequirePackage{algorithmic}
\usepackage{hyperref}

\usepackage{wrapfig}
\usepackage{booktabs} 
\usepackage{makecell} 
\usepackage{multirow} 
\usepackage{graphicx} 

\newcommand{\method}{ORCHID\xspace}
\newcommand{\methodft}{ORCHID-ft\xspace}
\newcommand{\methodACT}{ORCHID-ACT\xspace}
\newcommand{\methodACTft}{ORCHID-ACT-ft\xspace}
\newcommand{\pihl}{HL\xspace}
\newcommand{\pill}{LL\xspace}
\newcommand{\pihlphi}{HL\xspace}
\newcommand{\pillpsi}{LL\xspace}

\definecolor{darkgreen}{RGB}{0,100,0}   
\definecolor{green}{HTML}{EE82EE}

\title{Online Self-Training for Co-Adaptation in Hierarchical Diffusion Policies}

\author{%
  Clémence Grislain \quad Mathilde Kappel \quad Olivier Sigaud \quad Mohamed Chetouani \\
  ISIR,  Sorbonne Université, CNRS, Paris, France \\
  \texttt{\textit{surname}@isir.upmc.fr} \\
}

\begin{document}

\maketitle

\begin{abstract}
Hierarchical policies decompose language-conditioned long-horizon robotic manipulation into a high-level planner and a low-level controller. However, effective coordination between HL and LL requires that both components operate on compatible subgoal distributions. We propose \method, a self-training framework that enables stable online improvement of hierarchical diffusion policies by aligning planning and control through iterative refinement. By filtering policy samples via environment feedback, \method identifies trajectories where the planner and controller are jointly successful and distills them back into both modules via supervised learning.
This process induces a bidirectional co-adaptation: the planner grounds its subgoals in the actual reaching capabilities of the controller, while the controller specializes in the trajectory structures the planner produces. By relying on supervised distillation of filtered on-policy samples, \method avoids the instability typical of online hierarchical gradient-based RL training with diffusion models. On the CALVIN benchmark, \method allows a lightweight, initially weak model to outperform pure offline methods, including a Vision-Language-Action 
model twice its size. Code is available at: 
\url{https://github.com/clemgris/ORCHID.git}
\end{abstract}

\section{Introduction}

Learning robotic manipulation from language requires mapping multimodal inputs --~typically visual observations and natural language instructions~-- into continuous robot actions. Directly mapping these high-dimensional inputs to actions is particularly challenging in settings involving long-horizon and diverse tasks \citep{yao2025bridginglanguageactionsurvey}. While monolithic Vision-Language-Action (VLA) models have demonstrated impressive performance, they typically require extensive pre-training on massive datasets \citep{kawaharazuka2025vla}. To achieve similar task complexity without such overhead, hierarchical policies are a widely adopted approach \citep{bai2025unifiedunderstandingrobotmanipulation}, decomposing decision-making into a high-level planner (HL) and a low-level controller (LL). Through this decomposition, HL reasons over sparse subgoal representations to enable efficient long-horizon planning, while LL focuses on reaching each subgoal through fine-grained control. 

Prior work has explored various types of subgoals, including keypoints~\citep{chen2024simplehierarchicalplanningdiffusion, xian2023chaineddiffuser}, contact points \citep{wen2024ATM}, end-effector poses \citep{ma2024hierarchicaldiffusionpolicykinematicsaware}, and visual subgoals \citep{black2023Susie, zhang2024LDC, reuss2024MDT, liang2024skilldiffuserinterpretablehierarchicalplanning, xie2025ldp}, with diffusion models emerging as a particularly expressive planner for high-dimensional subgoal distributions \citep{rombach2022stablediffusion, liu2024sora}.

Despite their promise, hierarchical policies can be greatly bottlenecked by the interface between planning and control. Effective coordination between HL and LL requires that HL generate subgoals not only relevant to the task but also that LL can actually reach. Conversely, LL has to learn to generate successful trajectories conditioned on the specific plan structure of \pihl. We term this as the \emph{HL-LL coupling problem}. Prior work has addressed this problem through intermediate 'glue' modules that select plans fitted for the controller \citep{ajay2023HIP, hatch2024ghilglue, kang2025taksie}, or cross-level shared representations that couple planning and control within a common embedding space \citep{reuss2024MDT, zhang2024LDC, xie2025ldp}. Yet, the former introduces proxy models that increase inference overhead and training complexity, while the latter imposes a shared representation that must simultaneously satisfy the divergent requirements of planning and control. Moreover, these methods rely solely on offline training which may only partially resolve the coupling mismatch as the planner has no signal about whether its subgoals fall within the controller's actual reaching capacity. Fully closing this gap therefore calls for online environment interaction, where the planner can receive direct feedback on the reaching capabilities of LL. Yet this path remains largely unexplored for hierarchical manipulation policies. In fact, online training of hierarchical policies is notoriously unstable \citep{levy2019HRL, make2022HRL_survey_challenges, nachum2018dataefficienthierarchicalreinforcementlearning}, and diffusion-based planners compound this challenge through their multi-step stochastic denoising, which produces high-variance gradient estimates \citep{ma2025efficient}. As a result, most modern hierarchical methods for language-conditioned manipulation remain restricted to fixed human-annotated datasets \citep{bai2025unifiedunderstandingrobotmanipulation}.

To overcome these limitations, we propose {\bf \method} (\textbf{O}nline Self-T\textbf{R}aining for \textbf{C}o-adaptation in \textbf{Hi}erarchical \textbf{D}iffusion policies), a self-training framework for iterative improvement of hierarchical diffusion policies from environment feedback. Our approach is inspired by the recent success of self-training in large language models, where techniques such as STaR \citep{zelikman2022star}, SPIN \citep{chen2024selfplayfinetuningconvertsweak}, and ReST \citep{gulcehre2023reinforcedselftrainingrestlanguage} have demonstrated that models can bootstrap higher-level reasoning and performance by distilling their own filtered outputs. The key insight is that these methods do not require differentiating through the generative process, while matching or exceeding the performance of gradient-based RL \citep{havrilla2024teachinglargelanguagemodels}. It requires only the ability to sample candidate outputs and filter them by quality, a property shared by hierarchical diffusion-based policies under binary reward. \method organizes training into a self-reinforcing cycle. At each iteration, the current hierarchical policy is deployed to collect trajectories -- obtained via repeated sampling and filtered by binary task success -- where planner and controller succeed jointly; these are then distilled back into both HL and LL via supervised learning. By relying entirely on supervised learning, \method inherits the stability observed in language model self-distillation while enabling continuous improvement beyond the coverage of the initial human dataset.

Empirically, we show that \method better aligns the hierarchical components: HL generates more reachable subgoals, while LL simultaneously specializes in generating actions to complete the task, conditioned on the specific plans of HL. This alignment, along with the stable improvement loop, significantly improves hierarchical diffusion policies on both the Franka-3Blocks and CALVIN benchmarks, even in low-data regimes. Ultimately, we show that our method enables a lightweight model to exceed performance of prior offline methods trained both from scratch and on large-scale pre-training datasets (VLA). 

To summarize, our contributions are twofold:

$\bullet$ We propose \method, a self-training framework that mitigates the HL-LL coupling problem through iterative environment interaction, leveraging supervised distillation of jointly successful trajectories to stably improve both the diffusion planner and controller.

$\bullet$ We empirically demonstrate that \method induces a bidirectional co-adaptation between planner and controller, enabling consistent improvement of hierarchical diffusion policies even under restricted data regimes, on both the Franka-3Blocks and CALVIN benchmarks -- where a lightweight model exceeds performance of offline methods including a VLA twice its size.

\vspace{-3pt}
\section{Related Work}
\label{sec:related_work}

\textbf{Self-training for language model reasoning.}
Expert Iteration~\citep{Anthony2017ExpertIteration} first demonstrated that models can bootstrap performance by using policy-guided search to generate improved outputs, and then distilling these back into the policy via supervised learning. This principle has since proven highly effective in online fine-tuning of LLMs, where supervised distillation of filtered on-policy samples consistently matches or exceeds gradient-based RL fine-tuning while remaining more stable~\citep{zelikman2022star, gulcehre2023reinforcedselftrainingrestlanguage, chen2024selfplayfinetuningconvertsweak, guo2025deepseek, havrilla2024teachinglargelanguagemodels}. This property makes self-training particularly attractive for architectures where policy gradients are unreliable, such as diffusion-based hierarchical planners, which directly motivates the core update mechanism of \method.

\textbf{Hierarchical policies for language-conditioned manipulation}
While self-training offers a stable improvement mechanism, applying it to hierarchical policies first requires addressing the HL-LL coupling problem -- recognized implicitly across prior work, though rarely as a unified challenge. TaKSIE \citep{kang2025taksie} and HL-Glue \citep{hatch2024ghilglue} introduce an intermediate "glue" model to perform HL subgoal selection based on estimated task progress reflecting LL's preferences. HIP \citep{ajay2023HIP} uses an auxiliary model during training to regularize a diffusion HL toward LL's preferences, but this direct regularization can induce high variance and training instability. Other methods enforce coupling through joint representations, either via shared network layers (MDT \citep{reuss2024MDT}) or by planning directly in the visual embedding space of LL (LDC \citep{zhang2024LDC}; LDP \citep{xie2025ldp}). However, learning a representation space that simultaneously satisfies the divergent requirements of planning and control remains a significant challenge. Furthermore, all these methods rely on a single offline training stage. Fundamentally, without online interaction, the planner has no signal about whether its subgoals fall within the controller's actual reaching capacity, a gap that cannot be closed by dataset size alone. In contrast, our approach drives bidirectional co-adaptation between HL and LL through iterative on-policy refinement, without introducing auxiliary models, shared latent constraints, or gradient-based coordination losses.

\textbf{Fine-tuning diffusion-based and hierarchical policies with environment feedback}
Yet enabling stable improvement of hierarchical diffusion policies from environment feedback remains an open challenge: gradient-based fine-tuning of diffusion policies must propagate through multi-step stochastic denoising processes \citep{ren2024diffusionpolicypolicyoptimization}, which can lead to instability that is further exacerbated in hierarchical architectures \citep{levy2019HRL, make2022HRL_survey_challenges, nachum2018dataefficienthierarchicalreinforcementlearning}. Existing methods for fine-tuning diffusion models either require dense rewards \citep{fan2023dpokreinforcementlearningfinetuning, black2024ddpo} or rely on preference datasets \citep{wallace2023diffusionmodelalignmentusing, yang2024D3PO}, neither of which is typically available in language-conditioned manipulation. An alternative avenue relies on additional learned components beyond the policy itself -- world and/or reward models~\citep{chandra2025diwadiffusionpolicyadaptation, han2026enhancingpolicylearningworldaction, jain2025smoothseaskilledsailor} -- but these methods target flat diffusion policies and increase system complexity substantially. DAgger~\citep{ross2011Dagger}-based approaches such as DifNav \citep{shi2025daggerdiffusionnavigationdagger} instead require an expert oracle for online corrections, which is unavailable in our setting. In contrast, \method improves hierarchical diffusion policies directly from sparse binary environment feedback, requiring no additional learned components and no expert oracle.

\section{Problem Statement}
\label{sec:preliminaries}

\subsection{Goal-conditioned MDP}
 We consider a problem of multitask language-conditioned manipulation from visual observation. This problem can be formalized as a Goal Conditioned Partially Observable Markov Decision Process (GC-POMDP)~\citep{kaelbling1998pomdp}:
$
\mathcal{M} = (S, A, \mathcal{T}, \rho_0, \Omega, O, G, R),
$
where $S$ denotes the state space, with the initial state sampled from the distribution $\rho_0$. The agent receives partial observations, which are images $o=O(s) \in \Omega = \mathbb{R}^{3 \times H \times W}$, where $H$ and $W$ are the image height and width. The agent selects actions in the action space $A$ (such as joint configurations or \textit{6-DoF} end-effector poses) conditioned on the observation $O(s)$ and a textual goal $g \in G \subseteq \mathcal{V}^*$, where $\mathcal{V}$ is a vocabulary and $\mathcal{V}^*$ the set of all finite sequences over $\mathcal{V}$. The goal set $G$ can be grouped into subsets $G_l$ corresponding to different tasks $l \in L$. The transition function $\mathcal{T}$ governs the environment dynamics. The pair $(s_0, l)$ characterizes the multimodal task context and $(o_0=O(s_0), g \in G_l)$ is the corresponding agent observation. The binary reward function $R: \tau \times (s_0, l) \rightarrow \{0,1\}$ indicates whether the trajectory $\tau = (s_0, a_0, \dots, s_N, a_N)$ successfully completes task~$l$.

In this setting, the reinforcement learning (RL) objective is to find an optimal policy $\pi^*$ that maximizes the expected return $J(\pi)$. A full list of notations is provided in Appendix~\ref{app:notations}

\subsection{Hierarchical Policy and the HL--LL Coupling Problem}

We consider a hierarchical architecture where a high-level planner $\pi^{HL}$ generates a sequence of $M$ visual observation subgoals $\hat o_i \in \Omega$, that we call a \emph{plan} $\hat{\zeta} = \langle \hat{o}_1, \dots, \hat{o}_M\rangle$. This plan sequentially conditions a low-level controller $\pi^{LL}$ that produces robot actions. The joint objective is to find an optimal pair $\pi^* = (\pi^{HL*}, \pi^{LL*})$ that maximizes the expected return 
$$
J(\pi^{HL}_\phi, \pi^{LL}_\psi) = \mathbb{E}_{\substack{(s_0, l) \sim \rho_0 \times L \\ 
g \sim G_l \\ \hat{\zeta} \sim \pi^{HL}_\phi(\cdot \mid O(s_0), g) \\ \tau \sim \pi^{LL}_\psi(\cdot \mid \hat{\zeta}, O(s_0)).}} \left[ R(\tau,\, s_0,\, l) \right]
$$
However, optimizing this objective is non-trivial. Even if each component is individually well-trained, the system can fail due to a distributional mismatch at the HL--LL interface: $\pi^{HL}$ may generate plans that are task-relevant but lie outside the execution capability of $\pi^{LL}$, while $\pi^{LL}$ may be poorly specialized to the particular subgoal distribution induced by $\pi^{HL}$. This coordination failure --~which we term the \emph{HL-LL coupling mismatch}~-- is a well-documented challenge in hierarchical reinforcement learning \citep{levy2019HRL, nachum2018dataefficienthierarchicalreinforcementlearning, make2022HRL_survey_challenges}, and persists in offline hierarchical imitation learning when the two components are trained independently on fixed datasets \citep{ajay2023HIP, hatch2024ghilglue}.

\section{Method}
\label{sec:method}

\begin{figure*}[!t]
    \centering
    \includegraphics[width=1.\linewidth]{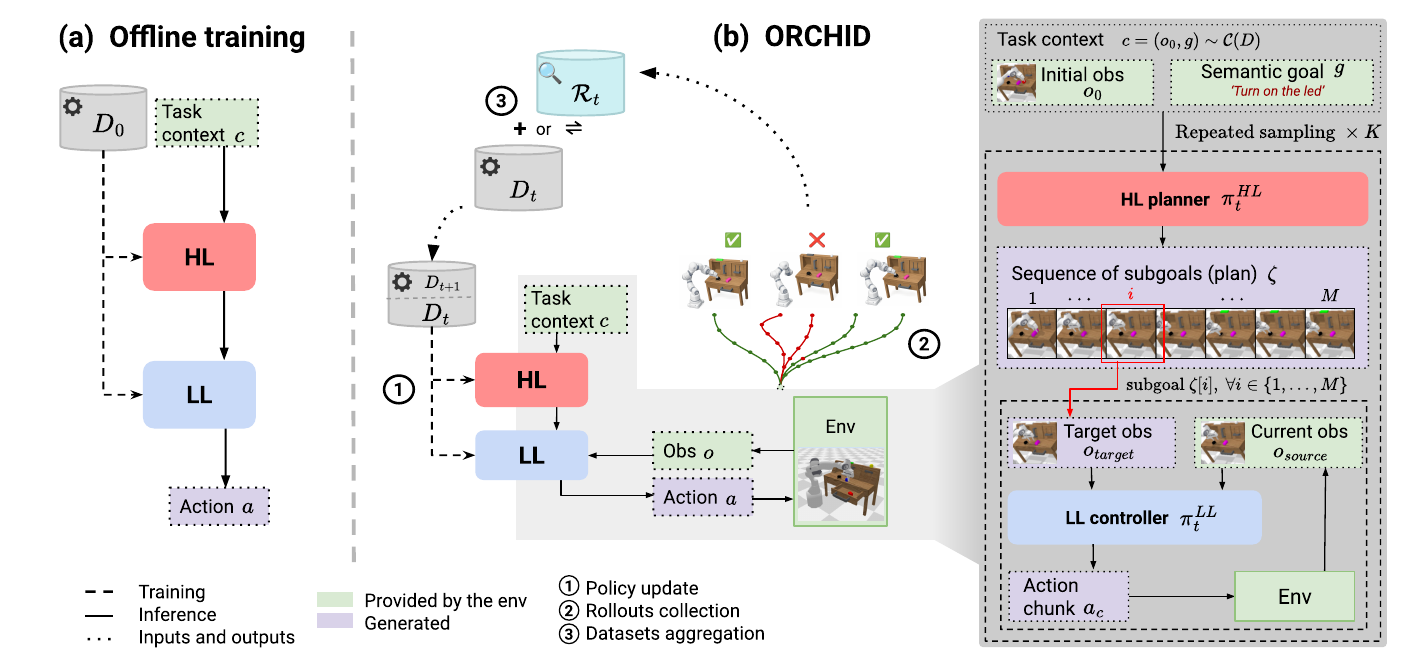}
    \captionof{figure}{\textbf{\method compared with common hierarchical policy training paradigms.} \textbf{(a)}: Independent training of HL and LL on a fixed offline dataset $D_0$, 
prone to HL-LL coupling mismatch. \textbf{(b)}: \method iteratively refines 
both components through: (1) supervised updates on $D_t$; (2) on-policy rollout 
collection filtered by sparse environment reward, producing $\mathcal{R}_t$ --- 
successful trajectories from which we extracts subgoals to train HL reachable by LL, and successful actions for LL conditioned on HL's plans; and 
(3) dataset aggregation.  \textit{Right:} During rollout collection, $\pi_{\text{HL}}$ generates $K$ subgoal sequences $\hat{\zeta}$ 
for each context in $\mathcal{C}(D_t)$, executed by $\pi_{\text{LL}}$ via 
action chunks $a_c$.}
\label{fig:main}
\end{figure*}

We address multitask language-conditioned manipulation with a hierarchical 
agent composed of a diffusion-based HL that generates 
a sequence of visual subgoals, and a controller LL that produces 
robot actions to execute them. As established in Section~\ref{sec:preliminaries}, 
maximizing $J$ requires simultaneously aligning HL plan generation with the actual capabilities of LL and aligning LL's actions with 
the specific plan structures produced by \pihl. Rather than relying solely on a fixed dataset $D_0$, Figure~\ref{fig:main}~(a), \method iteratively improves and aligns both components through a self-reinforcing loop, Figure~\ref{fig:main}~(b): 1) policy update with supervised training, 2) on-policy rollout collection filtered by environment reward, and 3) dataset aggregation. \method's pseudocode is given in Appendix~\ref{app:pseudo_code}.

\subsection{Hierarchical Diffusion Policy}
\label{subsec:hierarchical_policy}

\textbf{High-level planner.}
$\pi^{HL}_\phi$ is a diffusion model that takes the initial observation 
$o_0 = O(s_0)$ and textual goal $g$ as input, and outputs a visual plan 
$\hat{\zeta}$. Unlike prior work that generates subgoals reactively~\citep{black2023Susie, hatch2024ghilglue, 
reuss2024MDT, kang2025taksie}, we generate the entire plan at 
once~\citep{ko2023AVDC, du2023UniPi, zhang2024LDC}, invoking \pihlphi only 
at replanning. This reduces computational overhead and simplifies failure 
detection: if the agent fails to complete the plan, it triggers a full replan.

To train $\pi^{HL}_\phi$, from training trajectory $\tau^*$, we extract a sparse subgoal sequence 
$\zeta^* = \langle  O(s^*_{x_1}), \dots, O(s^*_{x_M})\rangle$.
\pihlphi is trained with the standard velocity-parameterized diffusion 
objective~\citep{karras2022elucidatingdesignspacediffusionbased}:
\begin{equation}
\mathcal{L}_{\mathrm{HL}}(\phi)
= \bigl\lVert v_\phi(\zeta^j,\, j,\, o_0^*,\, g) 
- (\alpha_j\,\epsilon_j - \beta_j\,\zeta^0) \bigr\rVert_2^2,
\label{eq:HL_loss}
\end{equation}
where $\zeta^j = \alpha_j\,\zeta^0 + \beta_j\,\epsilon_j$ is the noisy 
subgoal sequence at diffusion step $j$ and $\zeta^0 = \zeta^*$, and $\alpha_j, \beta_j$ characterize the noise scheduler.

\textbf{Low-level controller.}
$\pi^{LL}_\psi$ is a goal-conditioned visuomotor policy mapping a source 
observation $o_\text{source}$ and a target subgoal $o_\text{target}$ 
(generated by HL at test time) to an action chunk 
$a_c = \langle a_0, \dots, a_{n-1}\rangle$ that transitions the environment 
from $o_\text{source}$ to $o_\text{target}$. 

To train $\pi^{LL}_\psi$, from each training trajectory $\tau^*$, we sample 
contiguous chunks $a_c^* = \langle a_i^*, \dots, a_{i+m}^*\rangle$ with 
$m \sim \mathcal{U}[n_\text{min}, n-1]$, padding shorter chunks to fixed size 
$n$ with static actions. Variable chunk lengths improve robustness to varying difficulty of 
HL-generated subgoals by training $\pi^{LL}_\psi$ to reach subgoals at 
different temporal scales. Training minimizes
$$
\mathcal{L}_\text{LL}(\psi) 
= \bigl\lVert \pi^{LL}_\psi(o_\text{source},\, o_\text{target}) 
- a_c^* \bigr\rVert_2^2.
$$

\subsection{\method Training Loop}
\label{subsec:training}

\textbf{Stage 1 -- Policy Update} 

Both components are trained independently on the current dataset $D_t = \{(\tau, l)\}$ of successful observation-action trajectories, as described in Section~\ref{subsec:hierarchical_policy}. 

At $t{=}0$, components are randomly initialized and trained from scratch on the initial expert dataset $D_0$, corresponding to standard offline imitation learning (Figure~\ref{fig:main}~(a)); at $t{>}0$, $D_t$ is the dataset produced by data aggregation (Stage~3) of the previous iteration of \method~(Figure~\ref{fig:main}~(b)).

\textbf{Stage 2 -- Rollout Collection}

The current policy $\pi_t$ is then used to collect successful demonstrations by executing $K$ rollouts per context $(s_0, l) \in \mathcal{C}(D_t)$ and retaining the first success:
\[
\mathcal{R}_t = \bigcup_{(s_0,\, l)\,\in\,\mathcal{C}(D_t)} 
\Bigl\{\tau_{k^*} \;\Big|\; 
k^* = \min\bigl\{k \in [K] : R(\tau_k, s_0, l) = 1\bigr\}\Bigr\},
\]
with $o_0 = O(s_0)$, $g \sim G_l$. Contexts without a successful rollout 
contribute nothing to $\mathcal{R}_t$. To keep the dataset balanced, we cap the number of trajectories retained per task. $\mathcal{R}_t$  thus contains successful trajectories from which we extract states reached by LL as reachable subgoal targets for HL, and actions conditioned on HL's plans for LL.


A rich context set $\mathcal{C}(D_t)$ is critical for broad exploration. We use two complementary types: \textit{environment-reset} contexts ($O(s_0\sim \rho_{\text{reset}}),\, g \sim G_l$), which sample standard initial configurations, and \textit{replayed} contexts ($o_0 = o_N^*$ from $D_t,\, g \sim G_l$), which use terminal states of collected trajectories as new starting points. Since these terminal states feature object configurations that differ from standard resets, pairing them with sampled goals $g \sim G_l$ produces novel task contexts unreachable from $\rho_{\text{reset}}$ alone, expanding state coverage beyond $D_0$ without requiring an expert oracle.

\textbf{HL generating subgoals LL can reach.}
At $t=0$, $\pi^{HL}_\phi$ is trained on $\zeta^*$ extracted from offline expert trajectories -- subgoals that reflect the expert demonstration style (like human teleoperation) and can be unreachable for $\pi^{LL}_\psi$. At iteration $t>0$, the training targets $\zeta^0$ in Eq.~\ref{eq:HL_loss} are subgoals from $\mathcal{R}_t$: by construction, these are intermediate observations $O(s_{x_i})$ that $\pi^{LL}_\psi$ navigated through during successful rollouts. Training $\pi^{HL}_\phi$ to reproduce this distribution biases its generative distribution toward subgoals that $\pi^{LL}_\psi$ can actually reach.

\textbf{LL generating successful trajectories conditioned on HL plans.}
Simultaneously, \pillpsi is fine-tuned on trajectories conditioned on the specific plan structures of \pihlphi. This specializes the controller's action distribution to the subgoal patterns it will encounter at test time, reducing the distribution gap between the subgoals \pihlphi produces and those \pillpsi was trained on in $D_0$. Critically, both HL and LL updates are driven by the \emph{same} filtered dataset $\mathcal{R}_t$: as $\pi^{HL}_\phi$ is biased toward subgoals $\pi^{LL}_\psi$ can reach, $\pi^{LL}_\psi$ simultaneously specializes to the subgoal distribution $\pi^{HL}_\phi$ produces, inducing a bidirectional co-adaptation between planner and controller. By training only on these filtered successful rollouts, \method performs an implicit HL-LL alignment leading to a policy improvement step: by construction, this increases the likelihood of successful trajectories conditioned on HL's plans in the training distribution, which empirically translates to improved $J$ across iterations.

\textbf{Stage 3 -- Dataset Aggregation}

We consider two aggregation strategies:

\textbf{\method:} $D_{t+1} = D_t \cup \mathcal{R}_t$, followed by retraining from scratch. 
This approach prevents catastrophic forgetting and leverages all collected data, but incurs increasing computational cost as the dataset grows.

\textbf{\methodft:} $D_{t+1} = \mathcal{R}_t$, followed by fine-tuning from $\pi_t$. 
This results in constant computational cost at each iteration, at the risk of forgetting knowledge from previously collected data.

\section{Experimental Setup}
\label{sec:exp_setup}

\textbf{Architectures.} \method is applied to hierarchical diffusion (HD) policies based on a lightweight diffusion 3D CNN U-Net 
planner (AVDC-based~\citep{ko2023AVDC}) conditioned on a frozen CLIP embedding of the language goal. The low-level controller is either a Diffusion Policy (DP,~\citep{chi2024diffusionpolicyvisuomotorpolicy}, default) or an Action Chunk Transformer (ACT,~\citep{zhao2023learningfinegrainedbimanualmanipulation}), denoted \methodACT(-ft). Full architectural details are in Appendix~\ref{app:archi}.

\textbf{Metrics.} In order to evaluate to what extent HL generates plans that LL is able to reach, we introduce the \emph{reachability 
error} $\mathcal{E}$ as the expected observation-space distance between planned subgoals and observations of states actually reached by the 
controller:
$$
\mathcal{E}(\pi^{HL}_\phi, \pi^{LL}_\psi) = \mathbb{E}_{\substack{
(s_0, l) \sim \rho_0 \times L \\ 
\hat{\zeta} \sim \pi^{HL}_\phi(\cdot \mid O(s_0), g) \\ 
\tau \sim \pi^{LL}_\psi(\cdot \mid \hat{\zeta}, O(s_0))}} 
\left[ \frac{1}{M} \sum_{i=1}^{M} d\!\left(O(s_{x_i}),\, 
\hat{o}_i\right) \right]
$$
To ensure robustness to representation choice, we evaluate $\mathcal{E}$ across three embedding spaces using the $l2$ distance: pixel, R3M~\citep{nair2022r3m}, and DINOv2~\citep{oquab2024dinov2} (details in Appendix~\ref{app:metrics}).

Low $\mathcal{E}$ reflects the quality of the HL-LL interface but is insufficient for task success: visual subgoals lack full state information due to partial observability, so visual subgoal reachability does not guarantee task completion. We therefore also track the expected return $J$.


\textbf{Environments.} We evaluate \method on two benchmarks, further detailed in Appendix~\ref{app:envs}.

{Franka-3Blocks} comprises 10 language-conditioned manipulation tasks spanning pick-and-place, stacking, and pushing with $100$ demonstrations per task in $D_0$ in the default setting, and $10$ demonstrations per tasks in the low-data regime. Those demonstrations are collected from a hand-crafted expert with access to privileged environment and robot state information. 

{CALVIN}~\citep{mees2022calvin} contains 34 tasks with $150$ demonstrations per task in $D_0$ from human teleoperation; we use the $D{\rightarrow}D$ split. Despite being the in-distribution split, $D\rightarrow D$ is empirically the hardest (see CALVIN leaderboard \citep{mees2022calvin}) as it features the scarcest training dataset, making it the most stringent setting for evaluating gains from self-training under limited supervision. We evaluate our policies under with two protocols: single-task success in fixed settings out-of-distribution with respect to $D_0$ (\textit{MTLC}) and average length (Avg. Len.) of consecutive task completions starting from 1000 fixed reset distribution (\textit{LH-MTLC}, up to 5 tasks). In both protocols, evaluation uses unseen textual goals. MTLC success rate (SR) estimates $J$ on individual tasks, while Avg. Len. in LH-MTLC captures how well the policy sustains high $J$ across dependent task sequences which is a strictly harder criterion that penalizes error propagation over long horizons. 

\textbf{Baselines.} We compare \method (DP and ACT controllers, up to 3 or 4 iterations) against the HD policy trained on $D_0$ alone (iter~0) and five offline hierarchical baselines on CALVIN representing distinct strategies for coupling HL and LL: SuSIE~\citep{black2023Susie} (no HL-LL coupling), TaKSIE~\citep{kang2025taksie} (glue model), HULC~\citep{mees2022HULC}, MDT~\citep{reuss2024MDT}, and LDC~\citep{zhang2024LDC} (shared representations); see Figure~\ref{fig:main} and Appendix~\ref{app:baselines}. 

We further compare against FLOWER~\citep{reuss2025flower} which is the current SOTA on all CALVIN benchmarks. However, while the previous baselines are trained from scratch, FLOWER is a 950M parameter VLA that leverages a VLM pre-trained on internet-scale data, itself pre-trained on 250k robotic trajectories and then fine-tuned on the same initial dataset as the rest of the baselines. 

We restrict comparisons to offline baselines, as our goal is to improve policies beyond fixed offline datasets. Furthermore, to our knowledge, there is no prior work on fine-tuning hierarchical policies online on CALVIN. While online RL is a natural alternative, applying policy gradients to diffusion-based planners is unstable due to multi-step stochastic denoising, \method instead achieves environment-driven improvement through stable supervised self-training.

\sethlcolor{green!10}
\begin{figure*}[!ht]
  \centering
    \centering
    \captionof{table}{\small{\textbf{\method improves HD policies to achieve SOTA results on CALVIN LH-MTLC.} HD policies trained with different variants of our method (up to 4 iterations) against baselines on CALVIN LH-MTLC. We report the success rate for completing 1 to 5 consecutive tasks, alongside the average successful sequence length (Avg. Len.). Mean and standard error over 3 seeds. Best performance are highlighted in \textbf{bold}, while second best \underline{underlined}. Results from previous work are marked with $*$. $\dagger$ indicates training until convergence.}}
    \label{tab:big}
  \scriptsize
\resizebox{0.75\textwidth}{!}{\begin{tabular}{l l|ccccc|c}
\toprule
 & Method 
 & \multicolumn{5}{c|}{No. Instructions in a Row (1000 chains)} 
 & Avg. Len. ($\uparrow$) \\
\cmidrule(lr){3-7}
 &  & 1 & 2 & 3 & 4 & 5 &  \\
\midrule

\multirow{7}{*}{\rotatebox{90}{Baselines}}
& HULC*   & 82.7\% & 64.9\% & 50.4\% & 38.5\% & 28.3\% & 2.64 \tiny{($\pm$ 0.05)} \\
& SuSIE*  & 87.7\% & 67.4\% & 49.8\% & 41.9\% & 33.7\% & 2.80 \tiny{($\pm$ 0.15)} \\
& LDC*    & 88.7\% & 69.9\% & 54.5\% & 42.7\% & 32.2\% & 2.88 \tiny{($\pm$ 0.11)} \\
& TaKSIE* & 90.4\% & 73.9\% & 61.7\% & 51.2\% & 40.8\% & 3.18 \tiny{($\pm$ 0.02)} \\ 
& MDT*    & {93.7}\% & 84.5\% & 74.1\% & 64.4\% & 55.6\% & 3.72 \tiny{($\pm$ 0.05)} \\
& & & & & & &  \\
& FLOWER*    & 97.4\% & 92.4\% & \underline{86.9}\% & \underline{81.3}\% & \underline{74.9}\% & \underline{4.35} \tiny{($\pm$ 0.05)} \\

\midrule

\multirow{7}{*}{\rotatebox{90}{Ours}}
& \cellcolor{green!10} HD-ACT $\dagger$ (iter 0)
& \cellcolor{green!10} 76.0\% & \cellcolor{green!10} 49.3\% & \cellcolor{green!10} 31.4\% & \cellcolor{green!10} 19.9\% & \cellcolor{green!10} 12.3\% & \cellcolor{green!10} 1.89 \tiny{($\pm$ 0.02)} \\

& \methodACTft (iter 3)
& 87.0\% & 71.2\% & 56.2\% & 43.6\% & 31.8\% & 2.90 \tiny{($\pm$ 0.03)} \\


& & & & & & & \\

& \cellcolor{green!10} HD $\dagger$ (iter 0) 
& \cellcolor{green!10} 83.9\% & \cellcolor{green!10} 65.2\% & \cellcolor{green!10} 51.4\% & \cellcolor{green!10} 39.5\% & \cellcolor{green!10} 29.2\% & \cellcolor{green!10} 2.69 \tiny{($\pm$ 0.02)} \\

& \methodft (iter 3)
& 93.2\% & {85.4\%} & {76.6\%} & {66.9\%} & {57.3\%} & {3.80} \tiny{($\pm$ 0.01)} \\


& \method (iter 3)
& \underline{97.5\%} & \underline{92.7\%} & {86.6\%} & {79.3\%} & {71.3\%} & {4.28} \tiny{($\pm$ 0.03)} \\


& \method $\dagger$ (iter 4)
& \textbf{99.4\%} & \textbf{96.6\%} & \textbf{92.1\%} & \textbf{86.2\%} & \textbf{77.7\%} & \textbf{4.52} \tiny{($\pm$ 0.03)} \\

\bottomrule
\end{tabular}
}
\end{figure*}

\begin{figure}[!t]
\begin{minipage}[!t]{0.45\textwidth}
\centering
    \vspace{-10pt}
    \includegraphics[width=0.68\linewidth]{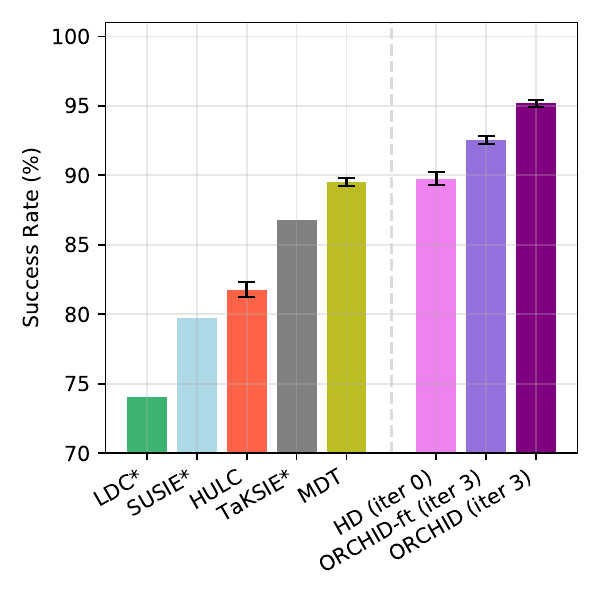}
    \vspace{-10pt}
    \caption{\small{\textbf{\method outperforms the hierarchical offline baselines on CALVIN MTLC.} Mean SR across tasks of the HD policy trained on $D_0$ and after 3 iterations of each version of \method compared to baselines ($*$ for results from previous work, bars indicates standard error over 3 seeds for ours and re-evaluated methods).}}
        \label{fig:main_mtlc}
\end{minipage}
\hfill
\begin{minipage}[!t]{0.53\textwidth}
    \centering
    \begin{subfigure}[!t]{0.48\textwidth}
        \centering
        \includegraphics[width=.95\linewidth]{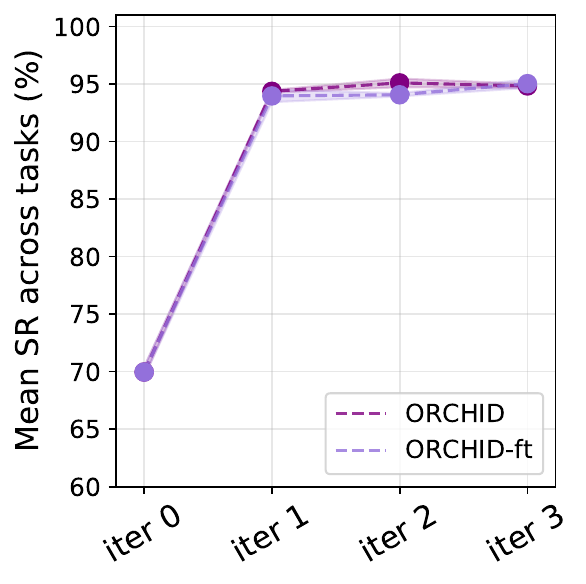}
        \label{fig:toy_env_sr}
    \end{subfigure}
    \hfill
    \begin{subfigure}[!t]{0.48\textwidth}
        \centering
        \includegraphics[width=0.95\linewidth]{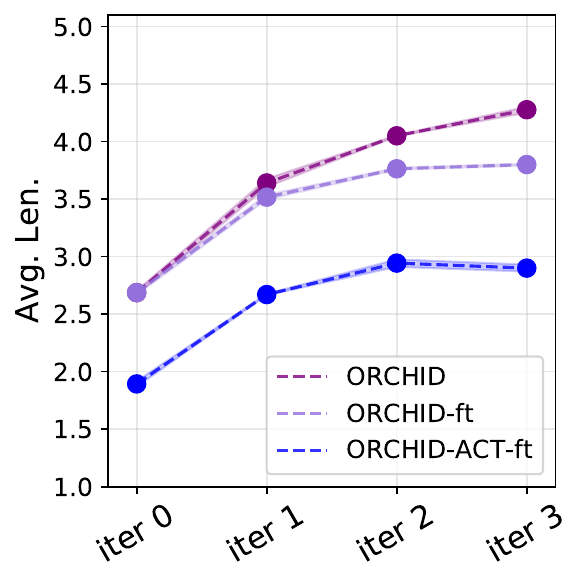}
        \label{fig:calvin_env_len}
    \end{subfigure}
\caption{\small{\textbf{Both variants of \method improve HD policies in Franka-3Blocks and CALVIN environments.} Performance of HD policies after 0 (offline training on $D_0$), up to 3 iterations of \method-(ft). Scatter for the mean metric and shaded areas for standard error across 3 seeds. (a) Mean SR across tasks in Franka-3Blocks. (b) Mean successful sequence length in LH-MTLC.}}
    \label{fig:improvement}
    \vspace{-6pt}
\end{minipage}
\end{figure}

\section{Results}
\label{sec:results}

\subsection{Main Results}
\label{sec:main_results}

Our self-training pipeline yields consistent gains across both benchmarks, improving the initial hierarchical policy after every iteration. We apply our iterative self-training for up to three iterations on Franka-3Blocks and CALVIN, and Figure~\ref{fig:improvement}(a,b) shows the resulting performance curves for both \method and \methodft.

On Franka-3Blocks, a single iteration increases the success rate from $70\%$ to over $94\%$.
On CALVIN, where task and semantic diversity are higher (Appendix~\ref{app:envs}), 
the gains are more gradual but remain monotonic.
In the fixed MTLC setting, \method improves success from $89.8\%$ to $95.2\%$, 
indicating generalization beyond the initial state coverage of $D_0$ (Figure~\ref{fig:main_mtlc}).
On LH-MTLC, it raises the mean number of successful tasks achieved in a row (Avg. Len.) from $2.69$ to $4.28$, 
more than doubling the success rate for 5 consecutive tasks 
($29.2\%\rightarrow71.3\%$ for DP with \method).
The largest gains consistently occur on the most difficult tasks (Appendix~\ref{app:taskwise}), 
highlighting and the benefits of \method for long-horizon problems.
Discussion on exploration can be found in Appendix~\ref{app:exploration} and qualitative failure cases in Appendix~\ref{app:failures}.
Critically, after three iterations, both \method and \methodft outperform all hierarchical baselines on both MTLC and LH-MTLC.

To push performance further, we train the best model with \method for a fourth iteration until convergence. Despite the fundamental difference in data and compute regimes (Appendix~\ref{app_subsec:computational_cost}), this exceeds performance in LH-MTLC, of FLOWER~\citep{reuss2025flower}, the current SOTA method on all the CALVIN benchmarks (Avg. Len. $4.52$ vs $4.35$, and 5-tasks SR $77.7\%$ vs $74.9\%$). This indicates that iterative self-training can close the gap to much larger offline-pre-trained models.


\textbf{Comparing LL architectures.}
Figure~\ref{fig:improvement} also reports the improvement curve for \methodACTft, which uses an ACT controller instead of a diffusion planner. The ACT-based policy starts lower, consistent with the stronger fit of diffusion controllers to multimodal action distributions, but \method still enables substantial iterative improvement, from $1.89$ to $2.90$ consecutive successful tasks on average in LH-MTLC which corresponds to an increase from $12.3\% \rightarrow 31.8\%$ for the success rate for 5 consecutive tasks (Table~\ref{tab:big}).

\textbf{Comparison of update strategies.}
On Franka-3Blocks, \method and \methodft perform similarly (Figure~\ref{fig:improvement}~(a)). On CALVIN, \methodft yields strong gains but tends to plateau, whereas \method continues to improve with further iterations (Table~\ref{tab:big}). This continued improvement comes at a higher computational cost, as discussed in Appendix~\ref{app_subsec:computational_cost}.

\begin{wrapfigure}{r}{0.45\textwidth} 
    \centering
    \vspace{-20pt} 
        \includegraphics[width=0.8\linewidth]{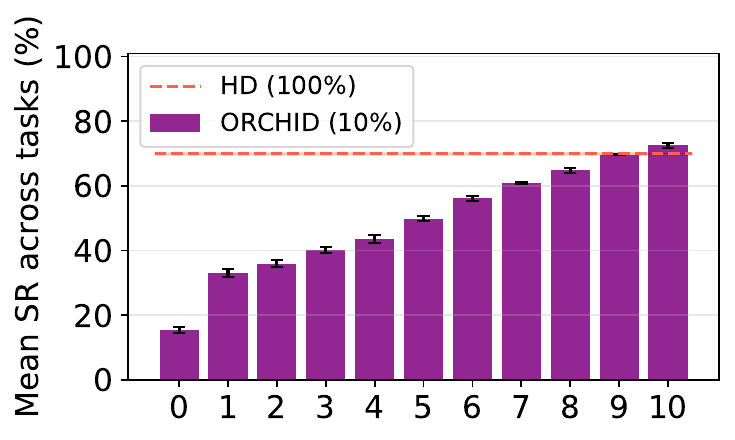}
        \vspace{-7pt}
        \caption{\small{\textbf{\method improves HD policy in low data regime in the Franka-3Blocks environment.} SR of HD policy after 0 (trained solely on 10\% of $D_0$) and up to 10 iter of \method. Mean SR and standard error across 3 seeds.}}
        \label{fig:low_data_regime}
        
    \vspace{-30pt} 
\end{wrapfigure}

\textbf{Low-data regime.}
We evaluate \method when the initial policy is trained on only 10\% of $D_0$ on Franka-3Blocks (10 demonstrations per task).
Under this scarce-data regime, Figure~\ref{fig:low_data_regime} shows that iterative self-training recovers the performance of the full-data baseline, improving the success rate from $15.6\%$ to $72.6\%$ after 10 iterations.
This demonstrates that \method is not contingent on a strong initialization and can bootstrap competent policies from minimal human supervision. Failure cases and task-wise analysis are further studied in Appendix~\ref{app:low_data}.   

\vspace{-7pt}
\subsection{Evidence for Bidirectional HL--LL Alignment}
\label{subsec:alignment}

\begin{wrapfigure}{r}{0.55\textwidth} 
    \centering
    \vspace{-13pt} 
\captionof{table}{
\textbf{Cross-evaluation of HL-LL combinations 
trained on expert data ($D_0$) or on-policy data ($R_1$)}, mean and standard errors over 3 seeds of task sequence length (Avg. Len.), 5-tasks SR and reachability error (RE) computed over all 
subgoals and the last subgoal, across pixel, R3M, and DINOv2 embedding spaces.}
\small{
\resizebox{0.55\textwidth}{!}{
\begin{tabular}{llcccc}
    \toprule
    
    & 
    & \cellcolor{green!10} $HL_{D_0}$ 
    & \cellcolor{blue!10} $HL_{R_1}$
    & \cellcolor{green!10} $HL_{D_0}$
    & \cellcolor{blue!10} $HL_{R_1}$ \\
    
    & 
    & \cellcolor{green!10} $LL_{D_0}$
    & \cellcolor{green!10} $LL_{D_0}$
    & \cellcolor{blue!10} $LL_{R_1}$
    & \cellcolor{blue!10} $LL_{R_1}$ \\
    
    \midrule
    \multicolumn{2}{l}{Avg. Len. \quad \multirow{2}{*}{($\uparrow$)}}
    & 2.69 \tiny${\pm0.01}$ 
    & 2.45 \tiny${\pm0.02}$ 
    & 2.86 \tiny{$\pm 0.01$}
    & \textbf{2.89 \tiny${\pm0.01}$} \\
        
    \multicolumn{2}{l}{5-tasks SR} 
    & 29.0\% \tiny${\pm0.7}$ 
    & 24.4\% \tiny${\pm0.4}$ 
    & 33.4\% \tiny${\pm 0.2}$
    & \textbf{34.7\% \tiny${\pm0.2}$} \\
    \midrule
    
    \multirow{2}{*}{RE-pixel ($\downarrow$)} 
    & All  
    & 12.30 \tiny${\pm 0.08}$ 
    & \textbf{11.42 \tiny${\pm 0.07}$} 
    & 12.76 \tiny${\pm 0.07}$
    & 12.56 \tiny${\pm 0.07}$ \\
    
    & Last 
    & 16.73 \tiny${\pm 0.26}$ 
    & 14.58 \tiny${\pm 0.27}$ 
    & 15.60 \tiny{$\pm 0.23$}
    & \textbf{12.80} \tiny${\pm 0.23}$ \\
    \cmidrule(lr){2-6}
    
    \multirow{2}{*}{RE-R3M ($\downarrow$) \tiny{(1e-3)}} 
    & All  
    & 4.78 \tiny${\pm 0.07}$ 
    & \textbf{4.33 \tiny${\pm 0.06}$} 
    & 4.90 \tiny{$\pm 0.06$}
    & 4.76 \tiny${\pm 0.07}$ \\
    
    & Last 
    & 6.84 \tiny${\pm 0.23}$ 
    & 6.17 \tiny${\pm 0.17}$ 
    & 6.33 \tiny{$\pm 0.21$}
    & \textbf{5.45 \tiny${\pm 0.23}$} \\
    \cmidrule(lr){2-6}
    
    \multirow{2}{*}{RE-DinoV2 ($\downarrow$)} 
    & All  
    & 7.80 \tiny${\pm 0.03}$ 
    & \textbf{7.47 \tiny${\pm 0.03}$} 
    & 7.94 \tiny{$\pm 0.03$}
    & 7.87 \tiny${\pm 0.03}$ \\
    
    & Last 
    & 9.37 \tiny${\pm 0.11}$ 
    & 8.60 \tiny${\pm 0.11}$ 
    & 8.95 \tiny{$\pm 0.10$}
    & \textbf{8.01 \tiny${\pm 0.10}$} \\
    \bottomrule
    
\end{tabular}
}
\label{fig:alignment_results}
}
\vspace{-15pt} 
\end{wrapfigure}

\textbf{Isolating the adaptation mechanism.}
\method induces co-adaptation through the self-training loop: because $R_1$ contains exclusively trajectories successfully executed by $\text{LL}_{D_0}$, training $\text{HL}$ on $R_1$ implicitly constrains its subgoal distribution to what the controller can actually reach. Conversely, training $\text{LL}$ on $R_1$ exposes it to the specific trajectory structures produced by the planner. To verify that both directions of this co-adaptation are necessary--and that the performance gains in Section~\ref{sec:main_results} stem from the on-policy structure of $R_1$ rather than increased data volume--we conduct a controlled ablation. Using matched dataset sizes ($|D_0|$=$|R_1|$), we compare the baseline policy $\pi_{D_0}$ against variants where components are replaced by $\text{HL}_{R_1}$ and $\text{LL}_{R_1}$, isolating each direction of adaptation.

\textbf{HL adapts to LL's capabilities.}
We first isolate planner adaptation by pairing $\text{HL}_{R_1}$ with the baseline controller $\text{LL}_{D_0}$. Since $R_1$ contains only trajectories that $\text{LL}_{D_0}$ successfully executed, $\text{HL}_{R_1}$ learns a subgoal distribution implicitly constrained by the LL's actual capabilities. Using $\text{HL}_{R_1}$ significantly lowers the reachability errors (Table~\ref{fig:alignment_results}) compared to $\pi_{D_0}$ ($p < 0.05$): HL learns to generate subgoals that are more consistent with what the controller can actually reach.

However, task success does not increase. This can be explained by the partial observability in the subgoal representation~\citep{chenu2023leveragingsequentialityreinforcementlearning}: a visual plan can correspond to various execution trajectories, some successful, some not. $LL_{D_0}$ has learned to generate trajectories (e.g. approach or grasp orientations) conditioned on human teleoperation subgoals, not on the distribution induced by $\text{HL}_{R_1}$, creating a distribution mismatch and ultimately poorer performance. This suggests that subgoal reachability is not sufficient; the controller must also adapt to the specific trajectory structures produced by the planner in order to translate subgoals reaching into task achievement.

\begin{wrapfigure}{r}{0.35\textwidth} 
    \centering
\captionof{table}{
Isolating LL performances trained on $\mathcal{D}_0$ and $R_1$ with GT subgoals. Mean SR on MTLC and standard errors over 3 seeds.}
\small{
\resizebox{0.31\textwidth}{!}{
\begin{tabular}{lcc} 
        \toprule
        & \multicolumn{2}{c}{GT} \\
        \cmidrule(lr){2-3} 
        & $LL_{D_0}$ & $LL_{R_1}$ \\
        \midrule
        SR ($\uparrow$) & 90.0 \tiny{$\pm0.1$} & 87.6 \tiny{$\pm0.1$} \\ 
        \bottomrule
    
\end{tabular}
}
\label{fig:LL_GT}
}
\vspace{-15pt} 
\end{wrapfigure}

\textbf{LL adapts to HL's guidance.} The failure of the mixed policy $(\text{HL}_{R_1}, \text{LL}_{D_0})$ implies that the controller must also adapt to the specific trajectory structures produced by the planner. To test this, we first pair $\text{HL}_{D_0}$ with $\text{LL}_{R_1}$: task success improves (+$0.17$ Avg. Len., +$4.4$pp 5-task SR) but reachability error on intermediate subgoals does not improve, since $\text{HL}_{D_0}$ still generates subgoals from the original offline distribution. We then evaluate the co-adapted policy $\pi_{R_1}$. 

This joint configuration achieves higher task success (+$0.20$ Avg. Len., +$5.7$pp 5-task SR, compared to $\pi_{\mathcal{D}_0}$, Table~\ref{fig:alignment_results}). Compared to $(\text{HL}_{D_0}, \text{LL}_{R_1})$, $\pi_{R_1}$ shows lower reachability error across all embedding spaces ($p < 0.01$ for last subgoal), since $\text{HL}_{R_1}$ generates more reachable plans than its counterpart trained on $\mathcal{D}_0$. Compared to $(\text{HL}_{R_1}, \text{LL}_{D_0})$, $\pi_{R_1}$ achieves higher task success, reflecting a behavioral shift in the LL: trained on $R_1$, the controller adapts to the trajectory distribution induced by the planner, improving task completion. Table~\ref{fig:LL_GT} further rules out the simpler explanation that $\text{LL}_{R_1}$ is unconditionally a better controller: when both controllers are paired with ground truth (GT) subgoals, $\text{LL}_{R_1}$ performs worse than $\text{LL}_{D_0}$, confirming that $\pi_{R_1}$'s gains are specific to the distribution of $\text{HL}_{R_1}$ rather than reflecting general improvement.

\textbf{Takeaway.}
These results establish the following:
(1) adapting HL to LL alone improves subgoal reachability $\mathcal{E}$ but not task 
success $J$;
(2) adapting LL to HL alone improves task success $J$ but not consistently reachability 
$\mathcal{E}$ (especially on intermediate subgoals);
(3) jointly adapting both components yields the largest gains in both $J$ and 
$\mathcal{E}$ (on last subgoals).
This pattern, observed under controlled data budget, is consistent with a 
bidirectional alignment process rather than improvements driven solely by additional 
data. This emerges naturally from the self-training loop without explicit architectural 
coupling.

\begin{wrapfigure}{r}{0.45\textwidth} 
    \centering
    \vspace{-13pt} 
    \captionof{table}{\small{Fixed \method LL (iter 3) guided by GT subgoals, and the respective HL without replanning. Results on MTLC.}}
    \label{tab:oracle}
    \resizebox{0.43\textwidth}{!}{
        \begin{tabular}{c|c|c}
        \toprule
        HL & LL & SR (\%) \\
        \midrule
        GT & \multirow{3}{*}{\method (iter 3)} & 91.2 ($\pm$ 0.5) \\
        \makecell{\method (iter 3) \\ w/o replan. } & & \textbf{92.1} ($\pm$ 0.3) \\
        \bottomrule
        \end{tabular}
    }
    \vspace{-10pt} 
\end{wrapfigure}

\textbf{Combining data volume and on-policy alignment.} 
When combined with increased data volume, this alignment enables the HL to generate plans that are both task-relevant and specifically tailored to the LL's capacities. To validate this, we compared the LL (after three iterations) when paired with its matched HL versus ground truth (GT) subgoals extracted from expert validation data. While GT subgoals are inherently task-relevant, they reflect human teleoperation dynamics rather than the specific capacities of the trained controller.
 As shown in Table~\ref{tab:oracle}, the \method policy (with replanning disabled for a fair comparison) outperforms the GT-guided baseline ($92.1\%$ vs. $91.2\%$ SR). This provides additional evidence that HL adapts its subgoal distribution to better match the capabilities of LL. Furthermore, Appendix~\ref{app:sr_re_iter} shows that RE decreases monotonically as the success rate increases across \method iterations,  reinforcing co-adaptation as a core driver of performance gains.

\vspace{-5pt}
\section{Conclusion}
\label{sec:conclusion}

In this work, we introduce \method, a self-training framework that improves hierarchical diffusion policies through iterative interaction with the environment. We highlight the HL-LL coupling problem as a fundamental limitation of offline hierarchical training and show empirically that \method reduces this mismatch by inducing a bidirectional co-adaptation between planner and controller, identified in our ablations as a core driver of performance gains. This yields consistent improvement across the Franka-3Blocks and CALVIN benchmarks, even under scarce supervision, ultimately enabling a lightweight, initially weak policy to surpass all offline hierarchical baselines and exceed FLOWER a 950M-parameter pre-trained VLA, and the current CALVIN SOTA.

Despite these results, \method has limitations worth addressing. Computational cost grows with dataset aggregation (Appendix~\ref{app_subsec:computational_cost}), motivating novelty-driven data selection to bound dataset growth. Self-training also stalls at near-zero initial success (Appendix~\ref{app:low_data}) calling for curriculum learning strategies. Addressing both would broaden \method's applicability, including to real-world settings where its reliance on binary success signals is a practical advantage: recent VLMs have shown promise as zero-shot success detectors~\citep{duan2024aha, grislain2026ifailsense}, providing precisely the supervision our framework needs (further discussion on real-world deployment considerations in Appendix~\ref{app:real_world}).

\section*{Acknowledgments}
This work used IDRIS HPC resources under the allocation 2025-[AD011015740R1] made by GENCI. It was supported by the European Commission’s Horizon Europe Framework Programme under grant No 101070381 (PILLAR-robots), and was partially funded by the French National Research Agency (ANR) under the OSTENSIVE projects (ANR-24-CE33-6907-01).

\bibliographystyle{unsrt}
\bibliography{bib}


\appendix

\newpage
\appendix

\onecolumn

\section{Notations}
\label{app:notations}

\begin{table}[h!]
\centering
\caption{List of Notations}
\label{tab:notations}
\resizebox{\textwidth}{!}{
\begin{tabular}{cl}
\hline
\textbf{Symbol} & \textbf{Description} \\
\midrule
$x \sim \mathcal{X}$      & $x$ sampled from a uniform distribution over the set $\mathcal{X}$ \\
$\mathcal{X}^*$ & Set of all finite sequences over $\mathcal{X}$ \\
\midrule
$S$      & Environment state space \\
$\rho_0 \in \mathcal{P}(S)$      & Initial state distribution \\
$\rho_\text{reset}\in \mathcal{P}(S)$      & Environment reset state distribution \\
$\Omega = \mathbb{R}^{3\times H \times W}$ & Observation space (pixel space) \\
$H\times W$ & Height and width of the pixel space \\
$o = O(s) \in \mathbb{R}^{3 \times H \times W}$ & Observation \\
$A$ & Action space \\
$\tau \in (S \times A)^*$ & State-action pairs trajectory \\
$\mathcal{V}$ & Vocabulary \\
$G \subset \mathcal{V}^*$ & Set of textual goals \\
$L$ & Set of tasks \\
$G_l \subset \mathcal{V}^*$ & Set of goals characterizing task $l$ \\
$g\in G_l$ & Textual goal characterizing a task $l$ \\
$\mathcal{T}: S \times A \rightarrow S$ & Environment transition function \\
$R: \tau \times (s_0, l) \rightarrow \{0,1\}$ & Environment success (binary) reward function \\
$T_\text{max} \in \mathbb{N}$ & Environment maximum number of steps allowed to complete a task \\
\midrule
$\pi^{HL}_\phi$ & High-level visual planner with parameters $\phi$\\
$\zeta = \{o_{1}, \dots, o_{M}\} \in \Omega^M$ & Sequence of subgoals \\
$\{x_1, \dots, x_M\}$ & Indexes of states used as subgoals in trajectory $\tau = \{s_0,a_0 \dots, s_N,a_N\}$ \\
$M \in \mathbb{N}$ & Number of subgoals in $\zeta$ \\
$\zeta[i] = o_{i} \in \Omega$ & $i$-th subgoal of the sequence $\zeta$ \\
$\zeta^j \in \Omega^M$ & noisy subgoal sequence at HL diffusion step $j$\\
$\alpha_j, \beta_j$ & $j$-th HL diffusion parameters \\
$\lambda$ & Classifier-free guidance parameter \\
\midrule
$\pi^{LL}_\psi$ & Low-level controller with parameters $\psi$ \\
$n \in \mathbb{N}$ & Size of the action chunk \\
$a_c \in A^n$ & Action chunk \\
\midrule
$\pi$ & Hierarchical policy composed of a high-level planner $\pi^{HL}$ and a low-level controller $\pi^{LL}$\\
$\pi_{-1}$ & A randomly initialized hierarchical policy\\
\midrule
$t$ & \method iteration step \\ 
$N_\text{iter} \in \mathbb{N}$ & Number of \method iterations \\
$D_t$ & Expert demonstrations dataset used to train the policy $\pi_t$\\
$\mathcal{R}_t$ & Exploration dataset collected from repeated sampling of the current policy $\hat \pi_{t}$\\
$\mathcal{C}(D)$ & Contexts extracted from dataset $D$ \\
$K \in \mathbb{N}$ & Number of sampling trials per context \\
$N_\text{data} \in \mathbb{N}$ & Maximum number of data in $\mathcal{R}$ per task \\
$\mathcal{E}$ & Reachability error between generated plans (subgoals) and attained states \\
$d$ & Distance in the observation space used to compute reachability errors \\
$J$ & Expected return \\
\bottomrule
\end{tabular}
}
\end{table}

\section{Architecture and hyperparameter details}
\label{app:archi}

\subsection{High-level planner:}
We consider a HL component which is a CNN 3D Unet based on AVDC~\citep{ko2023AVDC} architecture. At each diffusion step $j$, the model is conditioned on the initial observation by concatenating $o_0$ to the noisy observations $o_{i}^j$ for $ i\in\{1, \dots, M\}$ in the sequence to be denoised $\zeta^j$, and is conditioned on the textual goal $g$ by extracting goal features and injecting them into the downsampling, middle, and upsampling layers of the 3D CNN U-Net. The textual goal is embedded with CLIP~\citep{radford2021CLIP} which is frozen during training. The total number of trainable parameters of HL is 200M.

\textbf{Sampling and guidance:} We apply classifier-free guidance (CFG) with parameter $\lambda$ specifically on the goal conditioning. During rollout collection, we set $\lambda=3$ to favor exploration; for evaluation, we increase to $\lambda=5$ to reduce stochasticity. The main experiments utilize a DDPM~\citep{ho2020DDPM} scheduler with 100 steps. In Appendix~\ref{app:ddpm_vs_ddim}, we investigate the impact of using a DDIM~\citep{song2022DDIM} scheduler with 10, 30, 50, and 70 steps to reduce inference overhead.

\textbf{Subgoal extraction:} We utilize a fixed sequence of $M=8$ visual subgoals. For a trajectory $\tau = \{s_0, a_0, \dots, s_N, a_N\}$, we extract $M+1$ observations from states with indices $\{x_0, \dots, x_M\}$, where $x_0=0$ (initial) and $x_M=N$ (task completion). Intermediate indices $x_1 \dots x_{M-1}$ are sampled evenly along the temporal horizon. For training HL as described in Section~\ref{subsec:training}, $o_0=O(s_0=s_{x_0})$ is used for conditioning while the sequence $\zeta = \langle O(s_{x_1}), \dots, O(s_{x_M}=s_N) \rangle$ is used as supervision. 

\subsection{Low-level controller:} 
In our experiments, we consider two types of LL controllers: 
\begin{itemize}
    \item \textbf{A goal-conditioned diffusion policy (DP)}, based on the architecture introduced in \cite{chi2024diffusionpolicyvisuomotorpolicy}. This architecture uses 1D CNN layers and denoises an action chunk by injecting observation features into the denoising process through FiLM layers~\citep{perez2017filmv}. We extend this conditioning to also incorporate the target observation $o_{\text{target}}$. This is the default controller used in our method (250M trainable parameters). \\
    \item \textbf{A goal-conditioned Action Chunk Transformer (ACT)} from \cite{zhao2023learningfinegrainedbimanualmanipulation}, based on a ResNet18 vision encoder and transformer encoder-decoder (51M training parameters). In a similar way as with the DP controller, we extend the ACT architecture to integrate conditioning on both the current observation and a target observation (using the same vision encoder).
\end{itemize} 

\textbf{Action Chunking:} For both environments, the controller outputs a fixed-length action sequence of size $n=8$. To extract the ground-truth action chunks from the expert trajectories $\tau = \{s_0, a_0, \dots, s_N, a_N\}$ (as described in Section~\ref{subsec:hierarchical_policy}) we extract chunks of size $m$ sampled between $\frac{N}{8}$ and $8$ and apply static padding if $m < 8$. These parameters are chosen based on the target environments, where trajectories are capped at a maximum length of $N=64$ (i.e., $8 \times 8$ steps). Note that the action chunk is executed in the environment in an open-loop manner.

\subsection{\method}
\label{app:hyperparmeters}

\begin{figure}[h!]
    \centering
    \begin{minipage}[ht]{0.55\textwidth}
\textbf{Training:} For both HL and LL components, we first train on $D_0$ up to loss convergence to establish the initial number of gradient updates $N_0$ (this depends on the environment and the specific controller architecture as detailed in Table~\ref{tab:train_hyperparameters}). We then apply different schedules depending on the training strategy to find the number of gradient updates $N_t$ at iteration $t$:
\smallbreak
$\bullet$ \method (standard): We use a linear schedule $N_{t} = N_{t-1} + 0.5 N_0$ to accommodate the increasing size of the aggregated dataset. 

$\bullet$ \methodft (fine-tuning): We apply a fixed schedule of $N_{t} = 0.1 N_0$ for LL and $N_{t} = 0.2 N_0$ for \pihl, for each iteration.

\end{minipage}%
\hfill
\begin{minipage}[ht]{0.43\textwidth}
\captionof{table}{Training hyperparameters for both environments and the different component architectures.}
\label{tab:train_hyperparameters}
\resizebox{0.95\textwidth}{!}{
\renewcommand{\arraystretch}{1.3}
    \begin{tabular}{|l|l|c|c|}
\hline
\multicolumn{1}{|l|}{Environment} & Component & Batch size & $N_0$    \\ \midrule
\multirow{3}{*}{CALVIN}           & HL        & 16         & 5$e^{5}$ \\ \cline{2-4} 
                                  & LL ACT    & 128        & 5$e^{5}$ \\ \cline{2-4} 
                                  & LL DP     & 64         & 1$e^{6}$ \\ \midrule
\multirow{2}{*}{Franka-3Blocks}   & HL        & 64         & 1$e^{5}$ \\ \cline{2-4} 
                                  & LL DP     & 64         & 3$e^{5}$ \\ \midrule
\end{tabular}
}
    \end{minipage}
\end{figure}
\method (with a DP controller) on the CALVIN benchmark (with hyperparameter values mentioned in Table~\ref{tab:train_hyperparameters}) requires approximately 15 hours on eight NVIDIA H100 GPUs for 1$e^6$ gradient updates of \pihl, and 9 hours on a eight H100 GPU for 1$e^6$ updates of \pill. For both components, we use the Adam optimizer~\citep{kingma2017adammethodstochasticoptimization} with learning rates $1e^{-4}$.

\textbf{Rollouts collection:} For the data collection process described in Section~\ref{subsec:training}, we collect as many demonstrations per task as in the initial dataset $D_0$, that is a total of 150 demonstrations in CALVIN and 100 in Franka-3Blocks. Specifically, for CALVIN, we sample 100 demonstrations from replayed contexts and 50 demonstrations from environment-reset contexts per task at each iteration. We chose to sample more heavily from replayed contexts because they provide greater diversity, as demonstrated in Appendix~\ref{app:exploration}. For the Franka-3Blocks environment, we sample 50 demonstrations from replayed contexts and 50 from environment resets. For both environments, we allow $K=5$ trials per context. Notably, across all experiments, we were able to successfully collect the target amount of data for every task. However, when starting from weaker initial policies, we recommend increasing $K$ or reducing the per-iteration data collection target to ensure that the generative search remains computationally feasible. When the success rate of the initial policy is $p$, the probability of observing at least one success in $K$ trials is $1-(1-p)^K$, which provides a practical guideline for selecting $K$ according to one computational resources and data collection target.

\section{\method pseudo-code}
\label{app:pseudo_code}

Algorithm~\ref{alg:full} details the complete training procedure of our proposed method as depicted in \figurename~\ref{fig:main}. The specific update and data aggregation strategies of \method are highlighted in \textcolor{purple!80!black}{purple}, while those for \methodft are highlighted in \textcolor{blue}{blue}. This procedure encompasses the three phases described in Section~\ref{sec:method}: first, the supervised update of both components on the current dataset (Algorithm~\ref{alg:st}); second, the collection of rollouts from both environment-reset and replayed contexts. Algorithm~\ref{alg:traj_coll} details this collection process, which builds upon the hierarchical inference procedure described in Algorithm~\ref{alg:rollout} (see far right of \figurename~\ref{fig:main}). Note that data collection from these two context types is performed separately to ensure diverse demonstrations are captured. Finally, the newly collected data are aggregated with (or replace) the current training set to form the data for the next iteration.

\begin{algorithm}[ht]
  \caption{\method}
  \label{alg:full}
  \begin{algorithmic}[1]
    \STATE {\bfseries Input:} 
        $\pi_{-1}$: randomly initialized hierarchical agent \\
        $D_0$: initial expert demonstrations \\
        $N_\text{iter}$: number of \method iterations \\
        $N_\text{data}$: maximum number of trajectories to be collected per task at each \method iteration \\
        $K$: number of rollout trials per context \\
        $M$: number of subgoals per trajectory \\
        $n$: action chunk size \\
        $env$: environment \\
        $T_\text{max}$: maximum environment steps \\
        \colorbox{purple!10}{%
  \makebox[\dimexpr\linewidth-2\fboxsep][l]{%
  {$N_1$: number of gradient updates for training from scratch (HL and LL) (default)} }%
  } \\
        \colorbox{blue!10}{%
  \makebox[\dimexpr\linewidth-2\fboxsep][l]{%
 { $N_2 < N_1$: number of gradient updates for fine-tuning from previous policy (HL and LL) (FT)}
        }%
  }
    \medbreak
    \STATE $D \leftarrow D_0$, $\pi \leftarrow \pi_{-1}$
    
    \FOR{$t = 1$ {\bfseries to} $N_\text{iter}$}
        \STATE \colorbox{purple!10}{%
  \makebox[\dimexpr\linewidth-2\fboxsep][l]{%
    $\pi \leftarrow \text{SupervisedTraining}(\pi_{-1}, D, N_1)$
    \hfill $\triangleright$ Policy update from scratch
  }%
  }
        \STATE \colorbox{blue!10}{%
  \makebox[\dimexpr\linewidth-2\fboxsep][l]{%
  $\pi \leftarrow \text{SupervisedTraining}(\pi, D, N_2)$ \hfill $\triangleright$ {Policy fine-tuning}
          }%
  }
        \medbreak
    
        \STATE $\mathcal{R}_\text{exp-cxt} \leftarrow$ CollectTrajectories($\pi^{HL}, \pi^{LL}, \mathcal{C}(D), K, N_\text{data}, M, n, env, T_\text{max}$) \hfill $\triangleright$ replayed contexts
        \STATE $\mathcal{R}_\text{reset-cxt} \leftarrow $ CollectTrajectories($\pi^{HL}, \pi^{LL}, \rho_\text{reset}, K, N_\text{data}, M, n, env, T_\text{max}$) \hfill $\triangleright$ Environment-reset contexts
        
        \medbreak
        
        \STATE $R\leftarrow \mathcal{R}_\text{exp-cxt} \cup \mathcal{R}_\text{reset-cxt}$
        \
        \STATE \colorbox{purple!10}{%
  \makebox[\dimexpr\linewidth-2\fboxsep][l]{%
  $D \leftarrow D \cup \mathcal{R}$ \hfill $\triangleright$ {Dataset update: augment existing dataset}
          }%
  }
        \STATE \colorbox{blue!10}{%
  \makebox[\dimexpr\linewidth-2\fboxsep][l]{%
  $D \leftarrow \mathcal{R}$ \hfill $\triangleright$ {Dataset update (FT): replace dataset for fine-tuning}
  }%
  }
    \ENDFOR
  \end{algorithmic}   
\end{algorithm}

\begin{algorithm}[ht]
\caption{SupervisedTraining}
\label{alg:st}
\begin{algorithmic}[ht]
    \STATE \textbf{Input:} $\pi$, dataset $D$, number of gradient updates $N$
    \STATE Update both components $\pi_\text{HL}$ and $\pi_\text{LL}$ independently in a supervised fashion on $D$ as detailed in Section~\ref{subsec:training} for $N=(N_{HL}, N_{LL})$ gradient updates, respectively for \pihlphi and  \pillpsi.
\end{algorithmic}
\end{algorithm}

\begin{algorithm}[ht]
\caption{CollectTrajectories}
\label{alg:traj_coll}
\begin{algorithmic}[ht]
\STATE \textbf{Input:} $\pi^{HL}, \pi^{LL}, \mathcal{C}_\text{ctx}, K, N_\text{data}, M, n, env, T_\text{max}$, \textit{Rollout}
\STATE $\mathcal{R} \leftarrow \emptyset$
\FOR{$(s_0, l) \in \mathcal{C}_\text{ctx}$}
    \FOR{$k = 1$ {\bfseries to} $K$}
        \STATE $\tau =$ Rollout($\pi^{HL}, \pi^{LL}, o_0, g, M, n, env, T_\text{max}$) \hfill $\quad \triangleright$ Sample from current policy
        \IF{$R(\tau, s_0, l) = 1$ {\bfseries and} $Count(l, \mathcal{R}) < N_\text{data}$}
            \STATE $\mathcal{R} \leftarrow \mathcal{R} \cup \{\tau\}$ \hfill $\quad \triangleright$ Filter based on feedback 
            \STATE {\bfseries break} \hfill $\quad \triangleright$ Keep only one successful trajectory per context
        \ENDIF
    \ENDFOR
    \IF{$\forall l \in \mathcal{L}, Count(l, \mathcal{R}) \geq N_\text{data}$}
        \STATE {\bfseries break}
    \ENDIF

\ENDFOR
\STATE \textbf{return} $\mathcal{R}$
\end{algorithmic}
\end{algorithm}

\begin{algorithm}[ht]
  \caption{Rollout}
  \label{alg:rollout}
  \begin{algorithmic}
    \STATE {\bfseries Input:} $\pi^{HL}_\phi$, $\pi^{LL}_\psi$,  task context $(o_0, g)$, number of subgoals $M$, action chunk size $n$, environment $env$, maximum environment steps $T_\text{max}$
    \STATE $\tau \leftarrow \emptyset$
    \STATE $d \leftarrow 0$, $done \leftarrow \textbf{false}$, $x \leftarrow o_0$
    \REPEAT
    \STATE $\hat{\zeta} \leftarrow \pi^{HL}_\phi(x, g)$ \hfill $\quad \triangleright$ Sample visual plan
    \FOR{$i=1$ {\bfseries to} $M$}
        \STATE $a_c \leftarrow \pi^{LL}_\psi(x, \hat{\zeta}[i])$ \hfill $\quad \triangleright$ Sample action chunk
        \STATE $\tau \leftarrow \tau \cup \{x, a_c\}$
        \STATE $(x, done) \leftarrow env.\text{step}(a_c)$ \hfill $\quad \triangleright$ Update environment
        \STATE $d \leftarrow d + n$
        \IF{\textit{done}} 
            \STATE break
        \ENDIF
    \ENDFOR
    \UNTIL{$done = \textbf{true}$ or $d \ge T_\text{max}$}
    \RETURN $\tau$
  \end{algorithmic}
\end{algorithm}

\newpage

\section{Environments}
\label{app:envs}

\begin{figure}[!ht]
    \centering
    \begin{minipage}[!t]{0.57\textwidth}
        \centering
        \captionof{table}{Environment characteristics}
        \small
        \resizebox{\textwidth}{!}{
        \begin{tabular}{lcc}
            \toprule
            & \textbf{Franka-3Blocks} & \textbf{CALVIN} \\
            \midrule
            No. Tasks & 10 & 34 \\
            Evaluation textual goal & Seen & Unseen \\
            No. textual goals in training per task & 1 & 10 \\
            No. initial training data per task & 100 & 150 \\ 
            No. evaluation settings per tasks (MTLC) & 100 & 30 \\
            Initial settings sampling (MTLC) & $\rho_\text{reset}$ & fixed \\
            No. tasks to complete in a row (LH-MTLC) & - & 5 \\
            Initial settings sampling (LH-MTLC) & - &$\rho_\text{reset}$ \\
            \bottomrule
        \end{tabular}
        }
    \end{minipage}%
    \hfill
    \begin{minipage}[!t]{0.40\textwidth}
        \centering
        \begin{subfigure}[t]{0.48\textwidth}
            \centering
            \includegraphics[width=\linewidth]{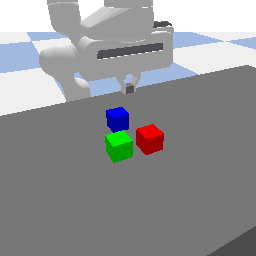}
            \caption{Franka-3Blocks environment}
            \label{fig:toy_env}
        \end{subfigure}
        \hfill
        \begin{subfigure}[t]{0.48\textwidth}
            \centering
            \includegraphics[width=\linewidth]{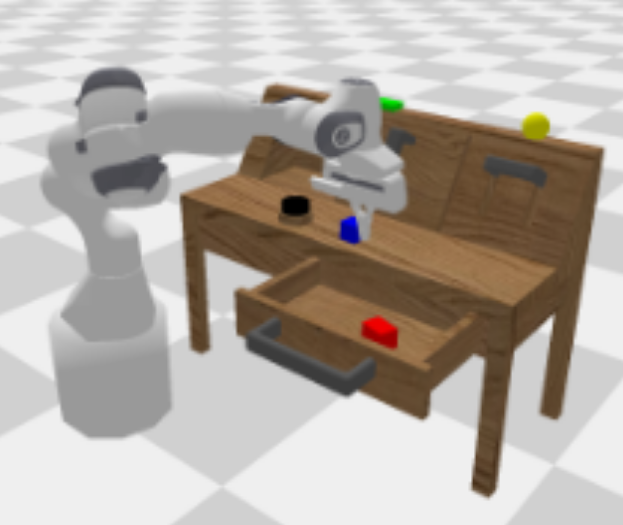}
            \caption{CALVIN environment}
            \label{fig:calvin_env}
        \end{subfigure}
        \caption{Environments}
        \label{fig:envs_summary}
    \end{minipage}
\end{figure}

\textbf{Franka-3Blocks environment:} As a proof of concept of the benefits of \method, we build a block manipulation environment based on the PyBullet physical simulator~\citep{pybullet}, which contains a Franka arm and three blocks (red, blue, and green) on a table, as shown in \figurename~\ref{fig:envs_summary}~(a). To create the initial demonstrations dataset $D_0$, we design a hard-coded expert relying on privileged information about the environment state (object positions and orientations, and the full robot state) and collect 100 demonstrations per task. We consider 10 tasks from three categories: lift tasks (3 tasks, one for each block); push tasks (6 tasks, pushing each block either \textit{right} or \textit{left}); and stack tasks (1 task for stacking any pair of blocks).
To characterize each task, we define a single language instruction per task, which is used for both training and evaluation, in order to reduce the complexity of generalizing to unseen textual goals.

For the low-data regime, we randomly sample 10 demonstrations per task from $D_0$ to form the new initial dataset.

\textbf{CALVIN environment:} To further evaluate \method on a more challenging setting, we test it on the two CALVIN Multi-Task Language-Conditioned (MTLC) and Long-Horizon Multi-Task Language-Conditioned (LH-MTLC) benchmarks. CALVIN introduces a multi-task setup where, unlike other environments such as Meta-World~\citep{yu2021metaworld} or RLBench~\citep{james2019rlbench}, almost every task can be executed in every setting. In other words, the setting does not induce the task, which is closer to real world multi-task problems.

\begin{figure}{r}
    \centering
    \vspace{-10pt} 

        \includegraphics[width=0.8\linewidth]{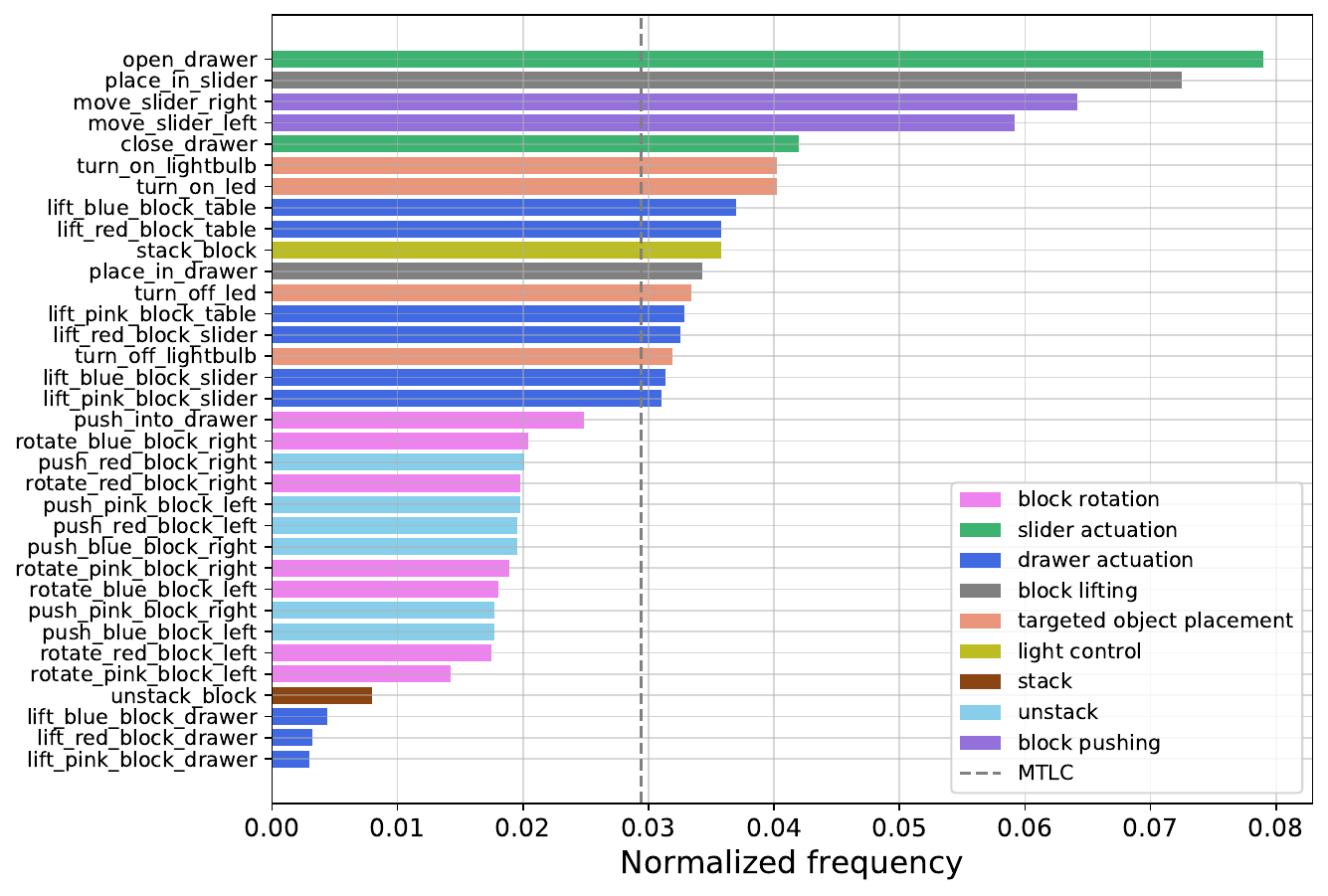}
        \caption{Task distributions in CALVIN LH-MTLC and MTLC (uniform).}
        \label{fig:task_dist}
\end{figure}
In the version we consider (task $D \rightarrow D$), the environment consists of a table with two lights (a \textit{LED} and a \textit{light bulb}) activated through different mechanical processes, a slider and a drawer. A Franka robot interacts with the table components as well as with three blocks of different shapes and colors (pink, red, and blue), as depicted in \figurename~\ref{fig:envs_summary}~(b). The environment includes 34 language-conditioned tasks, each associated with 11 textual goals: 10 for training and 1 for evaluation. These tasks can be categorized into: block spatial manipulation, slider actuation, drawer actuation, block lifting, targeted object placement, light control, stack and unstack. We assign a unique color to each category and plot the task distribution for both benchmarks, MTLC and LH-MTLC, in \figurename~\ref{fig:task_dist}. We observe that while the distribution is uniform in MTLC, LH-MTLC privileges certain categories over others due to the constraint of executing tasks sequentially.

To construct the initial dataset $D_0$, the benchmark provides 150 demonstrations per task collected via teleoperation, and 30 demonstrations per task for evaluation that are used as unseen initial settings for the MTLC benchmark. We further use these trajectories to extract ground truth subgoals in Section~\ref{sec:results}.

\section{Baselines details}
\label{app:baselines}
As mentioned in Section~\ref{sec:exp_setup}, we consider baselines belonging to three training paradigms of hierarchical policies for language-conditioned manipulation. We compare against these baselines on the CALVIN environment, using results from prior work when available or re-evaluating the baselines (using available checkpoints) when not (especially for MTLC). All these methods are trained solely on a fixed offline dataset and are categorized as follows:

\begin{wrapfigure}{r}{0.55\textwidth} 
    \centering
    \vspace{-10pt} 
\centering
    \includegraphics[width=\linewidth]{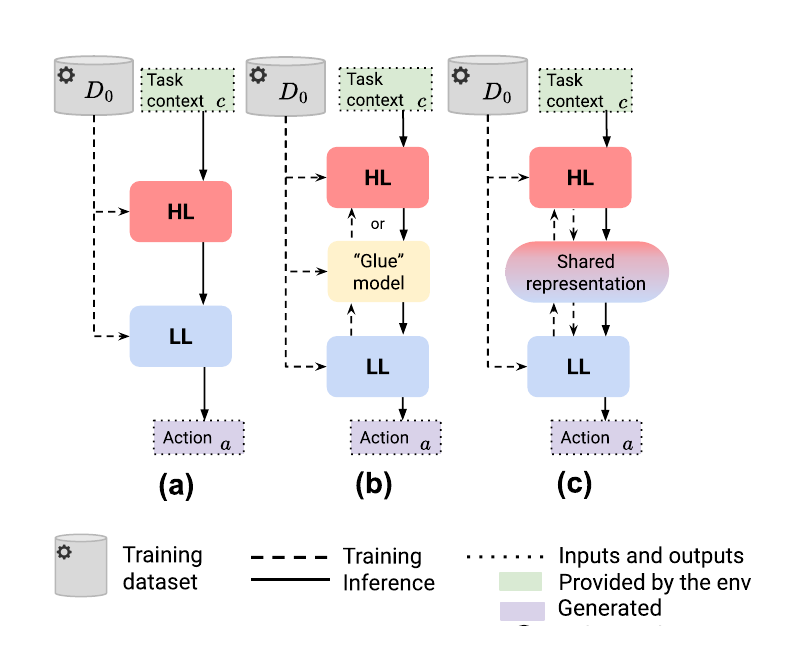}
    \caption{\textbf{Training paradigm of the offline hierarchical baselines} a) independent HL-LL training, b) 'glue' models c) cross-level representation space.}
    \label{fig:baselines}
    \vspace{-10pt} 
\end{wrapfigure}

$\bullet$ \textbf{Independent training, Figure~\ref{fig:baselines} (a):} Most existing hierarchical frameworks in robot learning train HL and LL independently, decoupling high-level planning from low-level control. In this work, we consider SuSIE~\citep{black2023Susie} as a representative of this category. This baseline is one of the first methods to use an image-editing model in a hierarchical policy for language-conditioned manipulation. SuSIE uses for HL a pre-trained text-to-image generative diffusion model, InstructPix2Pix~\citep{brooks2023instructpix2pixlearningfollowimage}, fine-tuned on robot video data, and a diffusion-based controller utilizing MLP layers and a ResNet-50 vision encoder. This method generates one subgoal at a time.

$\bullet$ \textbf{"Glue" models, Figure~\ref{fig:baselines} (b):} Methods have been proposed to enhance hierarchical diffusion policies through "glue" models that aim at bridging planning and control, hence mitigating the HL-LL coupling mismatch defined in Section~\ref{sec:preliminaries}. Among these methods, we use TaKSIE~\citep{kang2025taksie} as a baseline, which is based on SuSIE but improves it by incorporating a task-progress estimator fine-tuned from the pre-trained model LIV~\citep{ma2023livlanguageimagerepresentationsrewards} to select plans that help LL advance in the task. Except for the task progress estimator, this method shares the same characteristics as SuSIE.

$\bullet$ \textbf{Shared representations, Figure~\ref{fig:baselines} (c):} The final class of methods enforces coupling through shared cross-level representations. An early and widely used CALVIN baseline in this category is HULC~\citep{mees2022HULC}. In this method, the hierarchical policy is trained end-to-end with HL generating global discrete latent plans and LL learning local policies that are conditioned on these plans. LDC~\citep{zhang2024LDC} first trains a controller based on the HULC architecture (utilizing its discrete latent space) and then uses the trained latent space of this LL to generate plans using a diffusion \pihl. This way, HL generates compressed plans that contain exactly the information required by LL to solve the task. MDT~\citep{reuss2024MDT} is, at the time of writing, the current SOTA on the CALVIN ($D\rightarrow D$) LH-MTLC benchmark among hierarchical methods. In this method, HL is a transformer-based model that encodes the task context and current observation, leveraging a pre-trained Voltron~\citep{karamcheti2023voltron} vision-encoder, into a latent goal representation. This latent is then used to condition an attention-based diffusion \pill. The framework is trained using a joint objective: HL learns to represent the plan by predicting the next visual subgoal, while LL is trained via a diffusion loss to map these goal-conditioned latents to continuous actions. This ensures the latent space both encapsulates the necessary temporal information for planning and provides effective guidance for the low-level controller.

\section{Metrics}
\label{app:metrics}

In order to evaluate to what extent HL generates plans that LL is able to reach, we introduce the \textit{reachability error} defined in Section~\ref{sec:exp_setup}. Defining a robust distance metric $d$ over visual observations is non-trivial, as no single representation perfectly captures task-relevant configurations while remaining invariant to visual noise. To ensure a comprehensive assessment, we evaluate reachability error $\mathcal{E}$ using $l2$ distance across three complementary embedding spaces: \textbf{pixel-space} for raw scene sensitivity, \textbf{R3M}~\citep{nair2022r3m} for robotic-centric features learned from human manipulation, and \textbf{DINOv2}~\citep{oquab2024dinov2} for broad semantic and structural scene understanding. Reporting across these divergent metrics ensures that observed improvements reflect genuine physical reachability rather than artifacts of a specific representation. We report statistical significance using a one-sided Welch’s t-test over independent MTLC trajectories.


\section{Additional results}
\label{app:results}

\subsection{Main benchmarks}

Table~\ref{tab:mtlc_app} (MTLC) and Table~\ref{tab:big_app} (LH-MTLC) provide a detailed breakdown of the results presented in Section~\ref{sec:results}. We include performance metrics for the intermediate iterations of \method, \methodft, and \methodACT, as well as the ablation study using only environment-reset data collection (\methodft reset only), which is further discussed in Appendix~\ref{app:exploration}. All results are reported with standard error calculated over three seeds, along with those for the baselines.

{\centering
\captionof{table}{ Comparison of HD policies trained with different variants of our method (after 0 to 3 iterations) against baselines on CALVIN LH-MTLC. We report the percentage of trials successfully completing 1 to 5 consecutive instructions (mean over 3 seeds), alongside the average successful sequence length (Avg. Len.) and its standard error. Training until convergence is marked with $\dagger$. The best performance for each metric is highlighted in \textbf{bold}. Results from previous work are marked with $*$. }
\label{tab:big_app}
\resizebox{\textwidth}{!}{
\begin{tabular}{ll|ccccc|c}
    \toprule
& Method & \multicolumn{5}{c|}{No. Instructions in a Row (1000 chains)} &  \\
\cmidrule(lr){3-7}
  & & 1 & 2 & 3 & 4 & 5 & Avg. Len. ($\uparrow$) \\
    \midrule
    \multirow{5}{*}{\rotatebox{90}{Baselines}} 
 & HULC*      & 82.7\% & 64.9\% & 50.4\% & 38.5\% & 28.3\% & 2.64 ($\pm$ 0.05) \\
    & SuSIE*    & 87.7\% & 67.4\% & 49.8\% & 41.9\% & 33.7\% & 2.80 ($\pm$ 0.15) \\
    & LDC*      & 88.7\% & 69.9\% & 54.5\% & 42.7\% & 32.2\% & 2.88 ($\pm$ 0.11) \\
    & TaKSIE*  & 90.4\% & 73.9\% & 61.7\% & 51.2\% & 40.8\% & 3.18 ($\pm$ 0.02) \\ 
    & MDT*       & 93.7\% & 84.5\% & 74.1\% & 64.4\% & 55.6\% & 3.72 ($\pm$ 0.05) \\
    & FLOWER*    & 97.4\% & 92.4\% & {86.9}\% & {81.3}\% & {74.9}\% & {4.35} ($\pm$ 0.05) \\

    \midrule 
    
    \multirow{19}{*}{\rotatebox{90}{\textbf{Ours}}} 
    
    & HD-ACT$\dagger$ (iter 0)
      & 76.0\% \scriptsize{($\pm$ 0.3)} & 49.3\% \scriptsize{($\pm$ 0.7)} & 31.4\% \scriptsize{($\pm$ 1.0)} & 19.9\% \scriptsize{($\pm$ 0.6)} & 12.3\% \scriptsize{($\pm$ 0.4)} &  1.89 ($\pm$ 0.02) \\ 
             
    & & & & & & &\\ 
    & \methodACTft (iter 1)   & 85.3\% \scriptsize{($\pm$ 0.1)} & 66.0\% \scriptsize{($\pm$ 0.4)} & 50.8\% \scriptsize{($\pm$ 0.3)} & 37.8\% \scriptsize{($\pm$ 0.2)} & 27.0\% \scriptsize{($\pm$ 0.7)} & 2.67 ($\pm$ 0.01)       \\
    & \methodACTft (iter 2)    & 88.6\% \scriptsize{($\pm$ 0.8)} & 72.6\% \scriptsize{($\pm$ 0.7)} & 57.0\% \scriptsize{($\pm$ 0.8)} & 43.7\% \scriptsize{($\pm$ 0.4)} & 32.7\% \scriptsize{($\pm$ 0.6)} & 2.94 ($\pm$ 0.03) \\
    & \methodACTft (iter 3) & 87.0\% \scriptsize{($\pm$ 0.6)} & 71.2\% \scriptsize{($\pm$ 1.0)} & 56.2\% \scriptsize{($\pm$ 0.8)} & 43.6\% \scriptsize{($\pm$ 0.4)} & 31.8\% \scriptsize{($\pm$ 0.3)} & 2.90 ($\pm$ 0.03) \\
    & & & & & & &\\
    & HD$\dagger$ (iter 0) 
              & 83.9\% \scriptsize{($\pm$ 0.2)} & 65.2\% \scriptsize{($\pm$ 0.5)} & 51.4\% \scriptsize{($\pm$ 0.4)} & 39.5\% \scriptsize{($\pm$ 0.5)} & 29.2\% \scriptsize{($\pm$ 1.0)} & 2.69 ($\pm$ 0.02) \\ 
     & & & & & &\\ 
     & \methodft (iter 1)   & 91.2\% \scriptsize{($\pm$ 0.1)} & 79.9\% \scriptsize{($\pm$ 0.3)} & 69.8\% \scriptsize{($\pm$ 0.4)} & 59.9\% \scriptsize{($\pm$ 0.8)} & 50.7\% \scriptsize{($\pm$ 0.7)} & 3.52 ($\pm$ 0.02)       \\
    & \methodft (iter 2)    & 93.2\% \scriptsize{($\pm$ 0.3)} & 85.1\% \scriptsize{($\pm$ 0.1)} & 76.2\% \scriptsize{($\pm$ 0.2)} & 66.1\% \scriptsize{($\pm$ 0.5)} & 55.9\% \scriptsize{($\pm$ 0.5)} & 3.76 ($\pm$ 0.01) \\
    & \methodft (iter 3) &      93.2\% \scriptsize{($\pm$ 0.2)} &   85.4\% \scriptsize{($\pm$ 0.2)} &  76.6\% \scriptsize{($\pm$ 0.4)}  &  66.9\% \scriptsize{($\pm$ 0.2)}  &   57.3\% \scriptsize{($\pm$ 0.5)} & 3.80 ($\pm$ 0.01)\\
    & & & & & &\\ 
    & \methodft  (reset only) (iter 1)  & 88.5\% \scriptsize{($\pm$ 0.3)} & 68.8\% \scriptsize{($\pm$ 0.3)} & 47.7\% \scriptsize{($\pm$ 0.1)} & 29.3\% \scriptsize{($\pm$ 0.3)} & 15.1\% \scriptsize{($\pm$ 0.9)} &  2.50 ($\pm$ 0.02)       \\
    & & & & & & &\\ 
    & \method (iter 1) & 91.5\% \scriptsize{($\pm$ 0.4)} &  81.5\% \scriptsize{($\pm$ 0.7)} & 73.3\% \scriptsize{($\pm$ 0.8)} & 63.6\% \scriptsize{($\pm$ 1.0)} & 54.0\% \scriptsize{($\pm$ 0.4)} & 3.64 ($\pm$ 0.04)    \\
    & \method (iter 2) &    {95.1\%} \scriptsize{($\pm$ 0.3)}  &   {88.9\%} \scriptsize{($\pm$ 0.0)}   &  {82.0\%} \scriptsize{($\pm$ 0.3)} &   {74.2\%} \scriptsize{($\pm$ 0.2)} & {64.1\%} \scriptsize{($\pm$ 0.4)} &  {4.05} ($\pm$ 0.01) \\
    & \method (iter 3) & {97.5\%} \scriptsize{($\pm$ 1.0)} & {92.7\%} \scriptsize{($\pm$ 0.3)} & {86.6\%} \scriptsize{($\pm$ 0.5)} & {79.3\%} \scriptsize{($\pm$ 1.0)} & {71.3\%} \scriptsize{($\pm$ 1.0)} & {4.28} ($\pm$ 0.03)  \\
    & \method (iter 4) & {98.9\%} \scriptsize{($\pm$ 0.2)} & {95.2\%} \scriptsize{($\pm$ 0.1)} & {90.2\%} \scriptsize{($\pm$ 0.4)} & {83.5\%} \scriptsize{($\pm$ 0.3)} & {74.5\%} \scriptsize{($\pm$ 0.3)} & {4.42} ($\pm$ 0.01)  \\
    & \method $\dagger$ (iter 4) & \textbf{99.4\%} \scriptsize{($\pm$ .1)} & \textbf{96.6\%} \scriptsize{($\pm$ 0.2)} & \textbf{92.1\%} \scriptsize{($\pm$ 0.4)} & \textbf{86.2\%} \scriptsize{($\pm$ 0.5)} & \textbf{77.7\%} \scriptsize{($\pm$ .5)} & \textbf{4.52} ($\pm$ 0.02)  \\

    \bottomrule
  \end{tabular}
  }
}


\begin{wrapfigure}{r}{0.55\textwidth} 
    \centering
    \vspace{-10pt} 
    \captionof{table}{\textbf{Both versions of \method improve HD policies and achieve SOTA performances on CALVIN MTLC after only one iteration.} Mean success rate across tasks with standard error over 3 seeds of HD policies fine-tuned with 0 to 3 iterations of \method and \methodft compared with baselines. Best results in \textbf{bold}, second best \underline{underlined}, and results from prior work marked with $*$.}
    \label{tab:mtlc_app}
    \resizebox{0.40\textwidth}{!}{
    \begin{tabular}{ll|c}
        \toprule
         & Method & Success Rate ($\uparrow$) \\
         \midrule
         \multirow{5}{*}{\rotatebox{90}{Baselines}} & LDC* & 74.01 \% \\
         & SuSIE* & 79.73 \%\\
         & HULC & 81.77 ($\pm$ 0.55) \%  \\
         & TaKSIE* & 86.80 \% \\
         & MDT & 89.53 ($\pm$ 0.27) \%\\
         \midrule
         \multirow{9}{*}{\rotatebox{90}{\textbf{Ours}}} & HD (iter 0) & 89.77 ($\pm$ 0.46) \% \\
        \addlinespace[0.7ex]        
         & \methodft (iter 1) & 92.70 ($\pm$ 0.47) \% \\
         & \methodft (iter 2) & 91.87 ($\pm$ 0.27) \% \\
         & \methodft (iter 3) & 92.57 ($\pm$0.29) \%\\
         \addlinespace[0.7ex]
         & \method (iter 1) & 92.07 ($\pm$ 0.43) \% \\
         & \method (iter 2) & \underline{93.43} ($\pm$ 0.09) \%\\
         & \method (iter 3) & \textbf{95.20} ($\pm$ 0.25) \%\\
         \bottomrule
    \end{tabular}
    }
\end{wrapfigure}

\textbf{Baselines:} In both MTLC and LH-MTLC, the strongest baseline performance is achieved by TaKSIE~\citep{kang2025taksie} and MDT~\citep{reuss2024MDT}, both of which introduce mechanisms to bridge the HL-LL mismatch. Specifically, TaKSIE leverages a learned model of task progress to trigger replanning, while MDT introduces cross-level weights within the architecture to enforce coupling. Conversely, SuSIE~\citep{black2023Susie} performs among the worst in both benchmarks, as it relies on independent training of HL and \pill. On LH-MTLC, our initial architecture trained solely on $D_0$ achieves similar, though slightly lower, performance due to this same lack of alignment between planning and control leading to HL-LL mismatch. However, on MTLC, our initial model outperforms SuSIE; this may be attributed to our planner predicting an entire sequence of subgoals, thereby encapsulating both the spatial and temporal evolution of the environment, whereas SuSIE predicts only a single subgoal at a time.

\textbf{Intermediate iterations:} As shown in Table~\ref{tab:mtlc_app} and Table~\ref{tab:big_app}, the two update strategies--\method and \methodft--perform comparably after the first iteration. However, while \methodft plateaus after this initial stage on MTLC, \method demonstrates sustained improvement, ultimately reaching $95.20\%$. A similar trend is observed in Table~\ref{tab:big_app} for LH-MTLC: the performance of \methodft (using both ACT and DP controllers) plateaus after the second iteration, whereas \method exhibits consistent growth across all iterations. This phenomenon may be attributed to the limited training budget allocated to \methodft for computational efficiency (see Appendix~\ref{app:archi}); it is possible that increasing the number of gradient updates per iteration would mitigate this plateau. Consequently, the peak performances across both benchmarks are achieved by the third and the fourth iterations of \method.


\subsection{Exploration in \method}
\label{app:exploration}

\subsubsection{Exploration contexts}
\label{app:exploration_context}

\begin{wrapfigure}{r}{0.55\textwidth} 
    \centering
    \vspace{-10pt} 
    {\centering
    \captionof{table}{\textbf{Performance of the HD policy trained with one iteration of \methodft on CALVIN LH-MTLC.} Results report the mean and standard error across 3 seeds. Performance improvements relative to the HD policy trained on $D_0$ are marked in \textcolor{darkgreen}{green}, while degradations are marked in \textcolor{red}{red}. }
      \footnotesize
      \setlength{\tabcolsep}{2.5pt}
      \label{tab:reset_only}
      \resizebox{0.55\textwidth}{!}{\begin{tabular}{l|ccccc|c}
        \toprule
        Method & 1 & 2 & 3 & 4 & 5 & Avg. Len. ($\uparrow$) \\
        \midrule
        \methodft    & 88.5\% & 68.8\% & 47.7\% & 29.3\% & 15.1\% &  2.50 \tiny{($\pm$ 0.02)}    \\
        \makecell{(reset only) \\ (iter 1)} & 
        \textcolor{darkgreen}{($\times1.05$)} & 
        \textcolor{darkgreen}{($\times1.06$)} & 
        \textcolor{red}{($\times0.93$)} & 
        \textcolor{red}{($\times0.74$)} & 
        \textcolor{red}{($\times0.52$)} & \textcolor{red}{($\times0.93$)} \\
        \bottomrule
    \end{tabular}}
    }

\centering
    \includegraphics[width=0.9\linewidth]{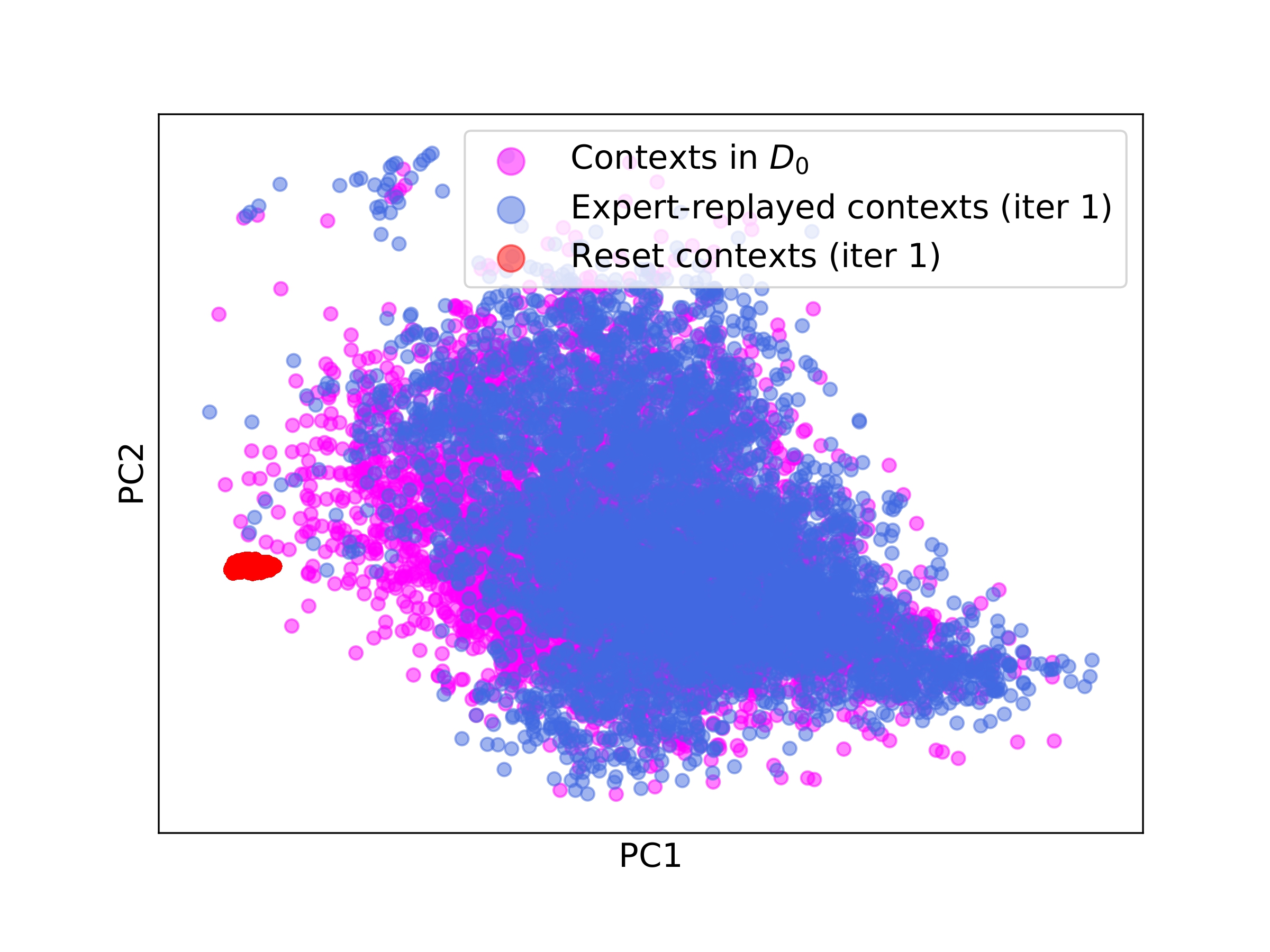}
    \caption{\textbf{replayed contexts increase the diversity of the context set, enhancing exploration.} PCA visualization of the environment initial states for contexts in $D_0$, in the replayed set extracted from $D_0$, and sampled from the environment reset state distribution.}
    \label{fig:context_dist}
\end{wrapfigure}

    To analyze the importance of augmenting the context set with replayed states, we compare \methodft in the LH-MTLC benchmark against a baseline where the exploration set of contexts $\mathcal{C}(D_t)$ is constructed solely from the environment reset distribution $\rho_{reset}$ (see Section~\ref{subsec:training}). To ensure a fair comparison, we maintain a constant number of sampled trajectories per task in both settings. The performance of the resulting hierarchical policy after one iteration is reported in Table~\ref{tab:reset_only}. As hypothesized, restricting collection to $\rho_{reset}$ reduces the diversity of the exploration dataset, which degrades overall performance. While this restricted approach yields some gains in short-horizon scenarios, where the state remains close to the reset distribution, performance sharply deteriorates when the task chain gets longer. Specifically, the success rate for completing 3 or more consecutive tasks falls below that of the initial policy, and is even almost divided by two ($\times0.52$) for completing 5 tasks in a row. This degradation is a direct consequence of the accumulating distributional shift inherent in long-horizon manipulation: as the agent executes sequential tasks, the environment state progressively diverges from the narrow support of $\rho_{reset}$. Without replayed contexts, the agent lacks the exploration required to handle these downstream states. Figure~\ref{fig:context_dist} further highlights the necessity of replayed contexts for robust exploration; it demonstrates that the diversity of reset contexts is highly limited compared to the broad state distribution provided by replayed trajectories, which is essential for the agent to generalize across long-horizon sequences.

\newpage

\subsubsection{State coverage}
\label{app:mode_collaps}

\begin{wrapfigure}{r}{0.35\textwidth} 
    \centering
    \vspace{-10pt} 
    {\centering
    \captionof{table}{\textbf{State coverage across \method iterations.} Convex hull area in PCA space of initial states from which the policy successfully completes a task, for each training dataset $\mathcal{D}_t$, from $t=0$ (offline only) to $t=3$.}
      \footnotesize
      \setlength{\tabcolsep}{2.5pt}
      \label{tab:state_coverage}
      \resizebox{0.35\textwidth}{!}{
      \begin{tabular}{lcccc}
        \toprule
        Training dataset & $\mathcal{D}_0$ & $\mathcal{D}_1$ & $\mathcal{D}_2$ & $\mathcal{D}_3$ \\
        \midrule
        Convex hull area & 79.7 & 85.6 & 90.7 & 94.5 \\
        \bottomrule
    \end{tabular}}
    }
\end{wrapfigure}

A known failure mode of iterative self-training \citep{gulcehre2023reinforcedselftrainingrestlanguage, chen2024selfplayfinetuningconvertsweak} is mode collapse in state coverage, where the policy stops expanding to new situations and keeps collecting data from the same initial conditions it has already mastered. To verify that \method does not suffer from this, we track the area of the convex hull of initial states and from which the policy successfully completes a task in PCA space, across iterations. The state features are the robot 7-DoF, scene state including object positions and orientations, light activation and slider/drawer opening extension. As shown in Table~\ref{tab:state_coverage}, the convex hull area of successful initial states increases monotonically across iterations, indicating hat each round of self-training expands the set of initial conditions from which the policy succeeds.

\subsection{Replanning}
\label{app:replanning}

\begin{wrapfigure}{r}{0.55\textwidth} 
    \centering
    \vspace{-15pt} 
\captionof{table}{Evaluation of HD policies with and without replanning when trained solely on $D_0$ (HD (iter~0) or after 3 iterations of \method. Results are reported as mean success rates and standard errors across 3 seeds on CALVIN MTLC.}
\centering
\label{tab:replanning}
    \small{
    \begin{tabular}{c|c}
    \toprule
    HL & SR (\%) \\
    \midrule
    
    \makecell{HD (iter 0) \\ w/o replan. } 
        & 83.3 ($\pm$ 0.1) \\
     & \\
    HD (iter 0)
        & 89.8 ($\pm$ 0.5) \\
    
    \midrule
    
    \makecell{\method (iter 3) \\ w/o replan. } 
        & 92.1 ($\pm$ 0.3) \\
        & \\
    \method (iter 3) 
        & 95.2 ($\pm$ 0.2) \\
    
    \bottomrule
    \end{tabular}
    }
\end{wrapfigure}

A central mechanism in hierarchical agents is the ability to replan when  \pillpsi fails to execute a plan or when \pihlphi generates task-irrelevant plans. This capacity is essential for robotic agents to handle execution errors, but it requires strong generalization from \pihlphi, as it must generate new plans from failed states. To assess the importance of this mechanism, we evaluate our HD policy trained solely on $D_0$, and after 3 iterations of \method with and without replanning on CALVIN MTLC. Table~\ref{tab:replanning} shows that the replanning mechanism provides a performance gain of $\sim 3\%$ to the final policy. This highlights the critical role of the replanning mechanism and demonstrates the robustness and generalization capabilities of our agent trained with \method, as it successfully recovers from states resulting from execution errors. 

Furthermore, we observe that the reliance on replanning decreases with our iterative refinement, as the initial HD policy performances dropped by $6\%$ with disabled replanning while the final policy only drop by $3\%$. This can be explained by two factors: \pihlphi has improved, generating more task-relevant plans due to bigger training set; and it has learn to generate plans that  \pillpsi can reach in a way that is successful for the task (Section~\ref{subsec:alignment}). Consequently, the planner generates higher-quality initial plans, reducing its dependency on the replanning mechanism.

\subsection{Task-wise analysis}
\label{app:taskwise}

\begin{figure}[h!]
\centering
\begin{minipage}[!t]{0.48\textwidth}

    \includegraphics[width=\linewidth]{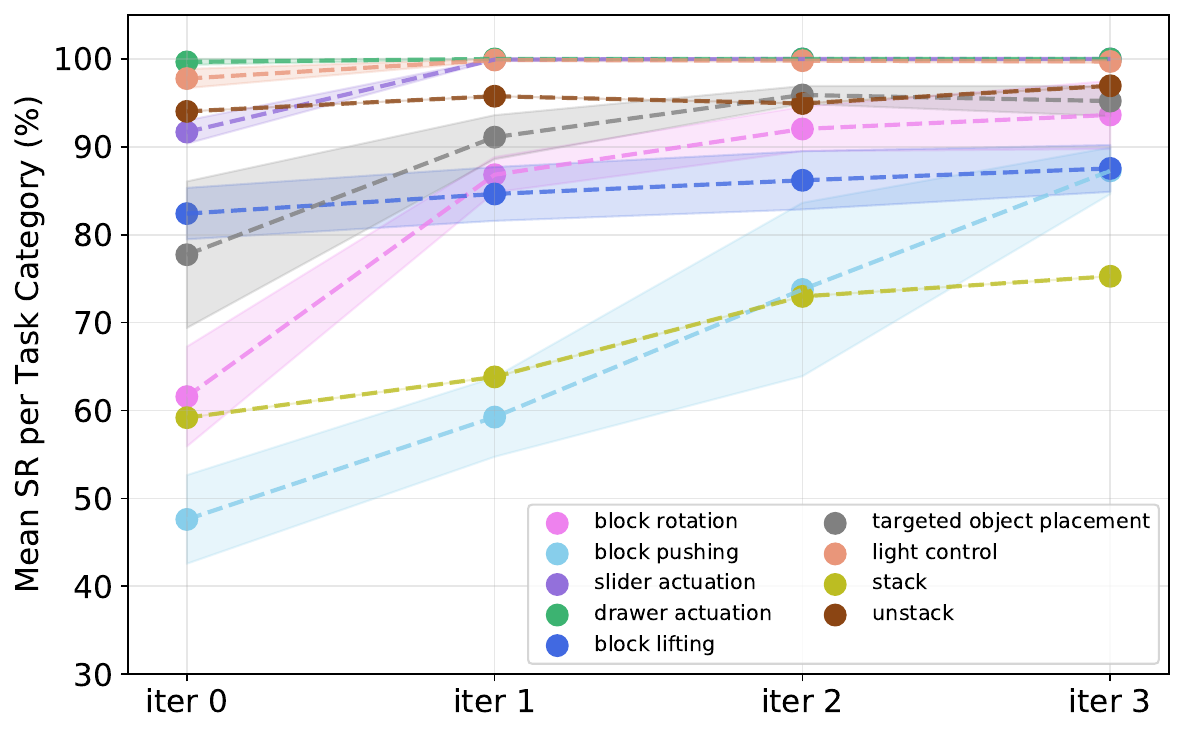}
\caption{\textbf{Iterative refinement yields the most significant improvements on complex tasks.} Mean success rates across task categories, with shaded regions indicating the standard error across tasks within each category, over 0-3 iterations of \method in the CALVIN LH-MTLC environment.}
    \label{fig:improvement_taskwise}
    
\end{minipage}
\hfill
\begin{minipage}[!t]{0.5\textwidth}
    \vspace{-25pt}
    \includegraphics[width=0.95\linewidth]{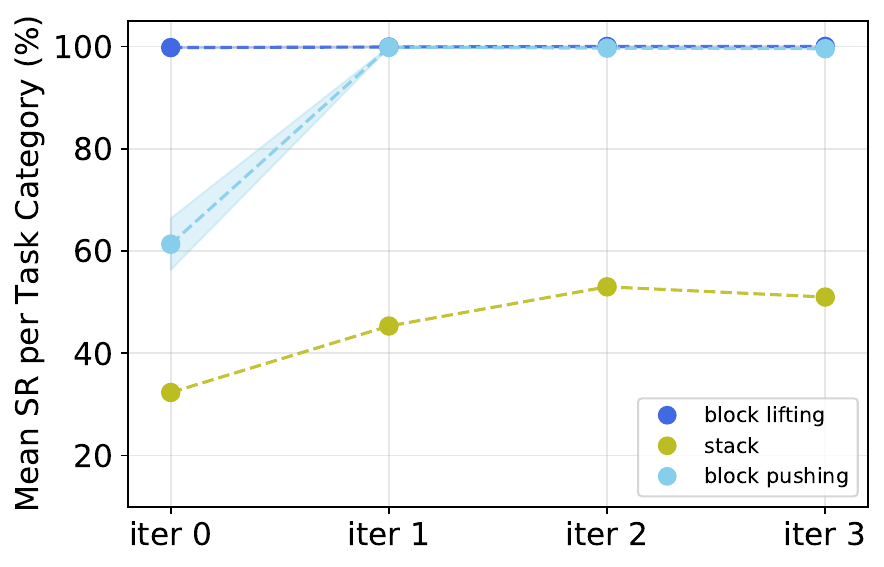}
\caption{\textbf{Mean success rates by task category in the Franka-3Blocks environment.} Results are shown over 0-3 iterations of \method, with shaded regions indicating the standard error across tasks within each category.}
\label{fig:improvement_taskwise_toy}
\end{minipage}
\end{figure}

Figure~\ref{fig:improvement_taskwise} illustrates the improvement in mean success rate across task categories in CALVIN (defined in Appendix~\ref{app:envs}) when training with \method, with similar results shown for the Franka-3Blocks environment in Figure~\ref{fig:improvement_taskwise_toy}. We observe that the success rate increases monotonically for all categories, demonstrating that our method positively impacts performance across the entire task distribution. This consistent growth is likely facilitated by our data collection strategy, which ensures a balanced training set by collecting an equal number of new demonstrations for every task. As shown in Figure~\ref{fig:improvement_taskwise}, after three iterations of \method, all task categories except \textit{stack} reach a mean success rate of at least $85\%$ in CALVIN. This consistent performance across tasks explains that the policies trained with our method perform similarly on both MTLC and LH-MTLC for achieving single tasks (see Table~\ref{tab:mtlc_app} and Table~\ref{tab:big_app}), even though the task distribution differs between the two benchmarks (Appendix~\ref{app:envs}). On both environments, the most significant gains are obtained in the \textit{block pushing} and \textit{block rotation} (for CALVIN) categories. These represent some of the most difficult tasks (alongside \textit{stack}) because they require the agent to ground textual goals into specific spatial movements.

\smallbreak

Qualitatively, we observe that the initial agent trained only on $D_0$ achieves between $50\%$ and $60\%$ success on these tasks, effectively choosing almost at random the correct movement direction (e.g., left vs. right). This phenomenon is further discussed in Appendix~\ref{app:failures}. Iterative refinement improves the grounding capabilities of the agent, mapping semantic instructions to the required physical movements. While the \textit{stack} category remains challenging due to its reliance on fine-grained control rather than high-level semantic understanding, \method still yields a significant performance increase of approximately $\times 1.25$ in CALVIN and $\times 1.43$ in Franka-3Blocks, over the initial hierarchical policy trained solely on $D_0$.

\subsection{Low-data regime}
\label{app:low_data}

\begin{wrapfigure}{r}{0.55\textwidth} 
    \centering
    \vspace{-10pt} 
\centering
    \includegraphics[width=0.9\linewidth]{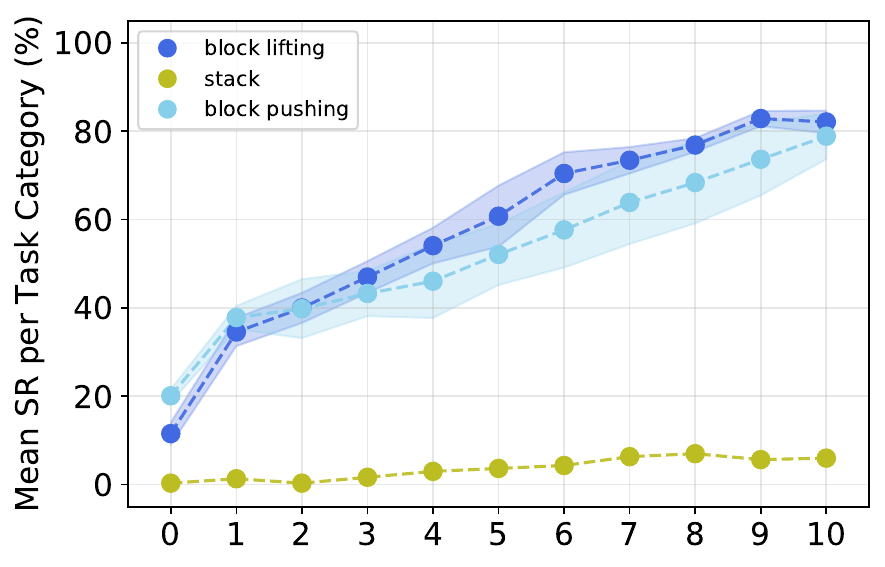}
    \caption{\textbf{Mean success rates by task category in the Franka-3Blocks environment starting from 10\% of $\mathcal{D}_0$.} Results are shown over 0-10 iterations of \method, with shaded regions indicating the standard error across tasks within each category.}
    \label{fig:toy_10_task_analysis_low_data}
\end{wrapfigure}

To evaluate how far \method can improve policies starting from weak initializations, we reduce the initial offline dataset to 10\% of $\mathcal{D}_0$ (10 demonstrations per task), yielding an initial policy with a mean SR of 15\% across tasks. We apply \method with $K=5$ and collect 15 trajectories per iteration (10 from the reset distribution and 5 from replayed contexts), for up to 10 iterations. To ensure balanced training across tasks, when the number of collected rollouts for a given task falls below the target, we oversample existing datapoints from that task so that each task contributes an equal number of training samples per iteration.

Figure~\ref{fig:toy_10_task_analysis_low_data} shows the per-category SR across iterations for \textit{block lifting}, \textit{block pushing}, and \textit{stack}. For tasks where the initial policy achieves non-negligible success ($20\%$ for \textit{block pushing} and $12\%$ for \textit{block lifting}), \method reliably bootstraps performance, reaching up to 80\% SR for both \textit{block lifting} and \textit{block pushing}. The \textit{stack} task, however, exposes a fundamental limitation shared with most reinforcement learning methods: when the initial policy achieves near-zero SR, there are no successful trajectories to filter and distill, stalling improvement entirely. This boundary condition is expected and consistent with the self-training framework's reliance on sparse but non-zero environment feedback as a learning signal.

\newpage

\subsection{Combining data volume and on-policy alignment}
\label{app:sr_re_iter}

\begin{wrapfigure}{r}{0.40\textwidth} 
    \centering
    \vspace{-10pt} 
\centering
    \includegraphics[width=\linewidth]{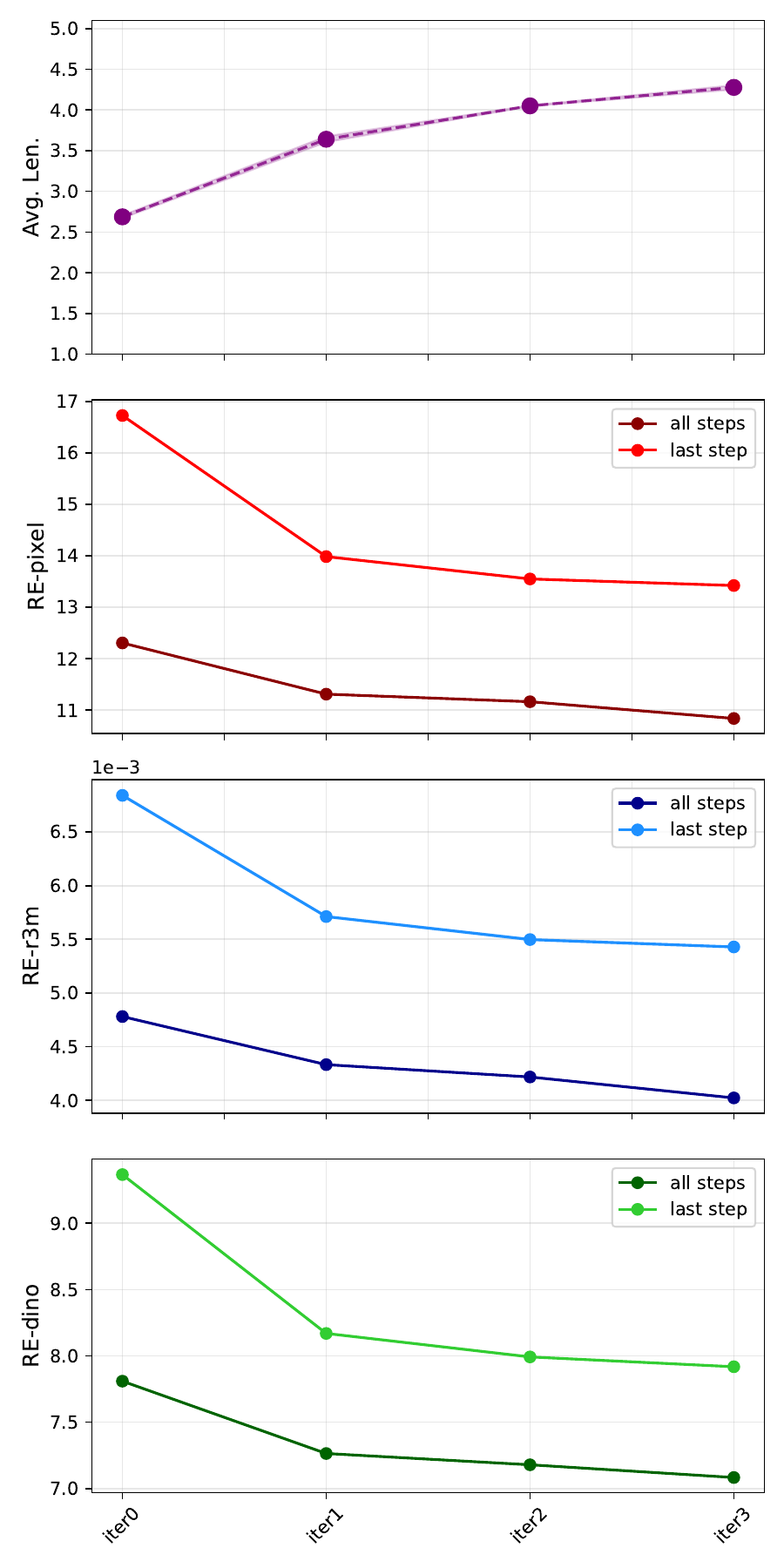}
    \caption{\textbf{All reachability errors decrease monotonically as performance increases across \method iterations.} The plot shows average successful sequence length (Avg. Len.) and reachability errors (all and last subgoals in pixel, R3M and DinoV2 embedding spaces) on LH-MTLC. Scatter points represent means, and shaded areas represent standard errors across 3 seeds.}
    \label{fig:sr_re_iter}
    \vspace{-200pt}
\end{wrapfigure}

We introduce the \textit{reachability error} $\mathcal{E}$ (Section~\ref{sec:exp_setup}) to measure how closely the observations reached by LL match the subgoals generated by HL. As established in Section~\ref{subsec:alignment}, low reachability error is not sufficient for high task success: it reflects the quality of the HL-LL interface, but visual subgoal proximity does not guarantee correct task execution. A reduction in $\mathcal{E}$ across iterations can stem from two complementary mechanisms: HL generating subgoals better aligned with LL's actual capabilities, and LL becoming more proficient at reaching the subgoals HL produces. Both are manifestations of the bidirectional co-adaptation \method induces.

Figure~\ref{fig:sr_re_iter} tracks $\mathcal{E}$ -- computed over all subgoals and the final subgoal, across pixel, R3M, and DINOv2 embedding spaces -- alongside average successful sequence length in CALVIN LH-MTLC, across iterations of \method. Reachability error decreases monotonically across all embedding spaces and both subgoal aggregations as task performance increases, throughout the iterations of \method. This consistent co-evolution of $\mathcal{E}$ and $J$ across iterations reinforces that the performance gains reported in Section~\ref{sec:main_results} are driven by progressive HL-LL co-adaptation, rather than by data accumulation alone.

\newpage

\subsection{Computational cost}
\label{app_subsec:computational_cost}

\subsubsection{Comparison between \method variants}

\begin{figure}[!h]
\centering
        \begin{minipage}[!t]{0.95\textwidth}
            \centering
            \begin{minipage}[!h]{0.55\textwidth}
        \begin{subfigure}[!h]{\textwidth}
            \includegraphics[width=0.75\linewidth]{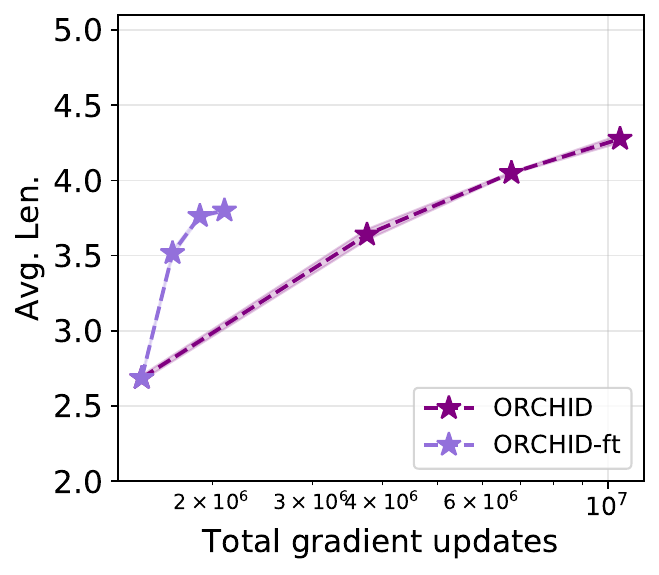}
            \caption{Average successful sequence length (Avg. Len.) in CALVIN LH-MTLC of policies trained with up to three iterations of \method and \methodft as a function of the required total number of gradient updates.}
            \label{fig:toy_env_len}
        \end{subfigure}
        \end{minipage}
        \hfill
        \begin{minipage}[!t]{0.43\textwidth}
        \begin{subfigure}[!t]{\textwidth}
            \vspace{-10pt}
            \includegraphics[width=0.95\linewidth]{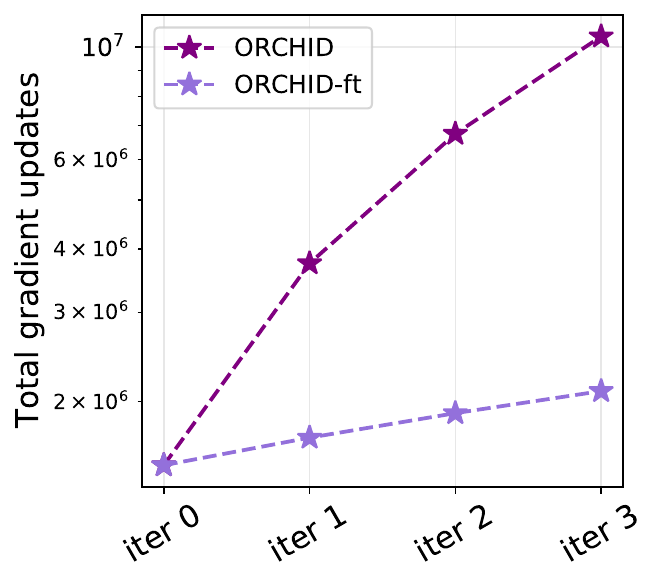}
            \caption{Total number of gradient updates required by iteration of \method and \methodft when training of the CALVIN environment.}
            \label{fig:calvin_env_updates}
        \end{subfigure}
        \end{minipage}
    \caption{\textbf{\method achieves better performance but requires more computation than \methodft.} Total number of gradient updates for training HL and LL (log scale) for \method and \methodft against performance and training iterations.}
        \label{fig:compute_cost}
    \end{minipage}
\end{figure}

A primary limitation of data aggregation is the increasing computational cost during training; as the dataset grows, the number of gradient updates required for convergence scales accordingly. Specifically, since \method retrains the hierarchical policy from scratch on all aggregated data, its total computational cost increases with the number of iterations. In order to mitigate this issue, we introduce \methodft, which fine-tunes the previous policy rather than starting from scratch and exhibits linear scaling, as it initializes the agent from the policy from the previous iteration (see Figure~\ref{fig:compute_cost}~(b)).
    
However, while \methodft ultimately reaches lower peak performance than \method on CALVIN, both variants behave similarly on the simpler Franka-3Blocks environment, despite \methodft being significantly less computationally costly. Notably, on CALVIN, three iterations of \methodft achieve performance similar to (and even slightly above) one iteration of \method while remaining less costly (see Figure~\ref{fig:compute_cost}~(a)). Further, when starting from a stronger initial policy, \methodft can be an effective solution for improvement with minimal computational cost. In this work, we used a small fraction of the initial number of gradient steps for training to limit costs, but this number could be increased to potentially mitigate the plateauing effect observed with \methodft (see \figurename~\ref{fig:improvement}). Yet, for greater improvement on more complex problems, the best performance is still achieved by \method.

\subsubsection{Comparison with offline methods.}
\method and offline methods such as FLOWER~\cite{reuss2025flower} and 
MDT~\cite{reuss2024MDT} operate under fundamentally different data and compute 
regimes. FLOWER is built on a 950M-parameter VLM backbone pre-trained on 
internet-scale vision-language data and fine-tuned on 250K robotic trajectories 
from diverse human teleoperation datasets before adaptation to the 5K CALVIN 
demonstrations. MDT similarly relies on a Voltron~\cite{karamcheti2023voltron} 
vision encoder pre-trained on large-scale video data for scene understanding. 
In contrast, \method starts from a randomly-initialized hierarchical diffusion policy  (450M parameters) trained on the same 5K CALVIN demonstrations, and improves it through iterative self-training by automatically collecting 20K additional 
trajectories over 4 iterations, corresponding to approximately 3M environment 
interaction steps, without additional human teleoperation or internet-scale 
pre-training.

Despite these different regimes, \method exceeds the performance of these methods 
on CALVIN. Rather than reflecting a direct comparison of data or compute 
efficiency, this result highlights a complementary axis of improvement: the gains 
from \method arise from targeted online refinement that reduces the HL--LL 
coupling mismatch, which remains difficult to address under purely offline 
training (see Section~\ref{sec:related_work}), together with the automatic on-policy data collection.

This comes with a distinct trade-off: \method relies on iterative environment 
interaction and dataset aggregation instead of large-scale offline pretraining, 
with computational cost growing with the number of iterations (or remaining 
bounded in the \methodft variant). These paradigms are therefore complementary rather than mutually exclusive.

\subsection{Failure analysis}
\label{app:failures}

\begin{figure}[!h]
    \centering
    \includegraphics[width=0.95\linewidth]{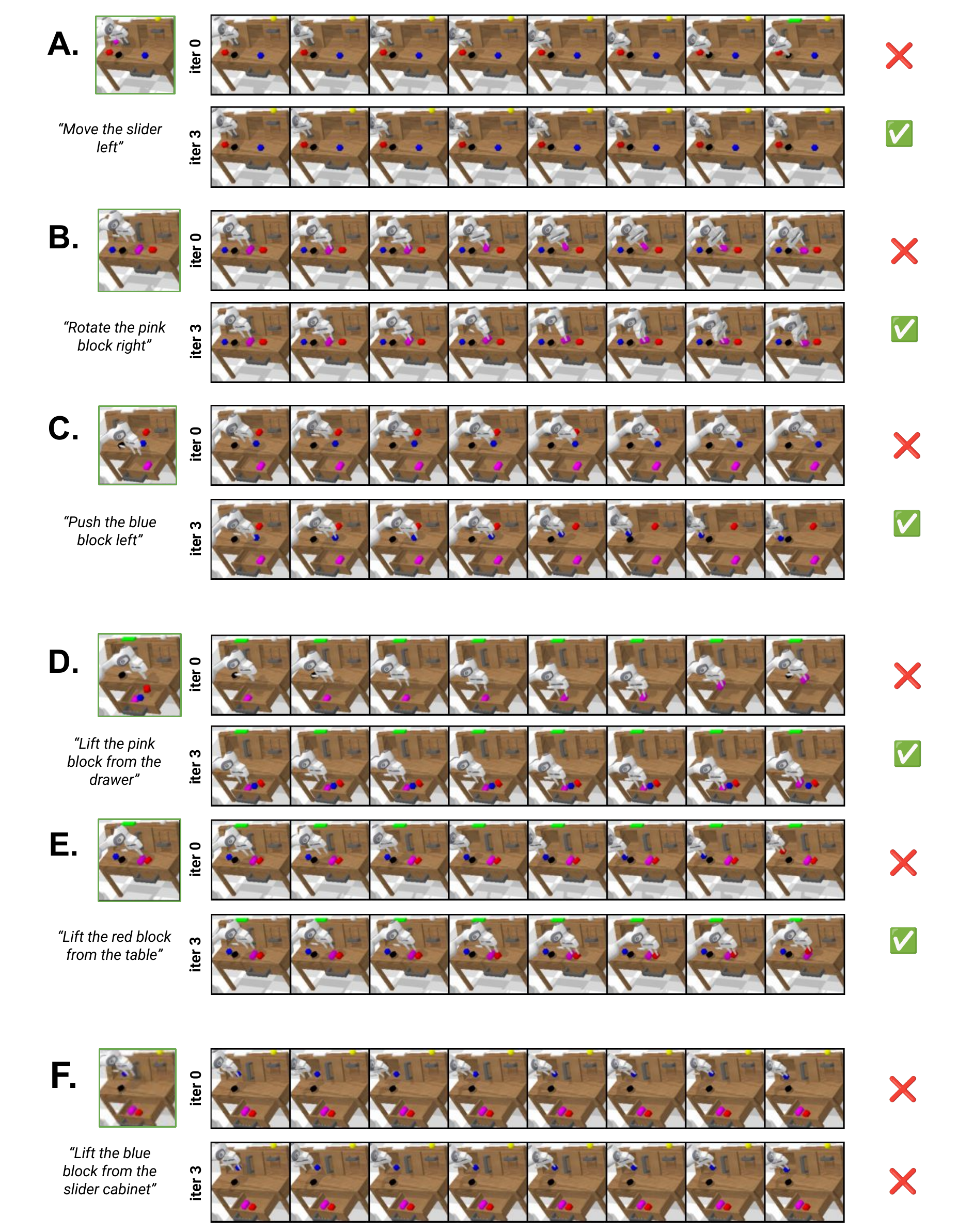}
\caption{\textbf{\method enables the generation of more task-relevant and feasible plans.} Qualitative analyses comparing plans generated by the HD policy trained solely on $D_0$ (iter 0) versus plans generated after three iterations of \method (iter 3) for same validation contexts in CALVIN MTLC. The symbol \textcolor{darkgreen}{\textit{check}} \faCheckCircle \vspace{1pt} indicates that the plan led to success, while the symbol \textcolor{red}{\textit{cross}} \faTimesCircle  \vspace{1pt} indicates failure.}
    \label{fig:failures}
\end{figure}

In order to provide a qualitative analysis of the improvement in plan generation to support the results in Section~\ref{sec:results}, we collected plans generated before and after applying \method for three iterations. These plans were generated for the same context extracted from the validation set of CALVIN MTLC and indicate whether they led to success or failure. Figure~\ref{fig:failures} illustrates these examples. Failure cases A, B, and C demonstrate that when trained solely on $D_0$, HL generates plans that are not always task-relevant. Failure case A can be attributed to the limited coverage of $D_0$. In fact, the agent might never have seen an initial state for the task 'move the slider left' where the slider is already positioned almost to the left. As a consequence, HL generated a plan that ignores the task and where the agent achieves another task (likely the task whose initial state distribution in $D_0$ is closest to the example initial observation, in this case 'Turn on the led'). Failure cases B and C can be explained by the lack of grounding capabilities of the initial agent; in both cases, the plans depict the robot interacting with the right object but performing the task incorrectly (pushing right for B and rotating left for C). For all these examples, the policies trained with \method (iter 3) avoid these errors and generate plans that align with the task and ultimately lead to success.

Examples D and E demonstrate that the initial planner can generate plans that are task-relevant but not consistent with either the initial observation of the environment (D) or with the physics of the world (E). In D, the red and blue blocks, while present in the initial observation, are no longer present in the generated plan, while in E, the plan depicts the robot moving toward the blue block, which transforms into a red block in the last generated image (likely to match the initial instruction). These errors are common in generative diffusion models as the generation is not constrained by physical rules. In contrast, the plans generated by the policy trained with \method demonstrate consistency with the initial observation and a higher consistency with physical rules, leading to higher success rates. Qualitatively, we observe that the remaining failures of the fine-tuned HD policy are due to fine-grained control, as shown in Example F, where both the initial and the fine-tuned planners generate a task-relevant and feasible plan but fail due to control errors.

\newpage

\newpage

\section{Deployment of \method : inference time and reward function design}
\label{app:real_world}

While this work evaluates \method in simulation, we discuss here two practical considerations for real-world deployment: inference speed and reward function design. We show that inference overhead from diffusion-based planning can be substantially reduced without performance loss, and that \method's reliance on binary success signals -- rather than dense rewards -- makes it compatible with recent VLM-based success detectors, lowering the barrier to automated real-world training.

\subsection{Reducing inference overhead}
\label{app:ddpm_vs_ddim}

\begin{wrapfigure}{r}{0.45\textwidth} 
    \centering
    \vspace{-10pt} 
        \includegraphics[width=\linewidth]{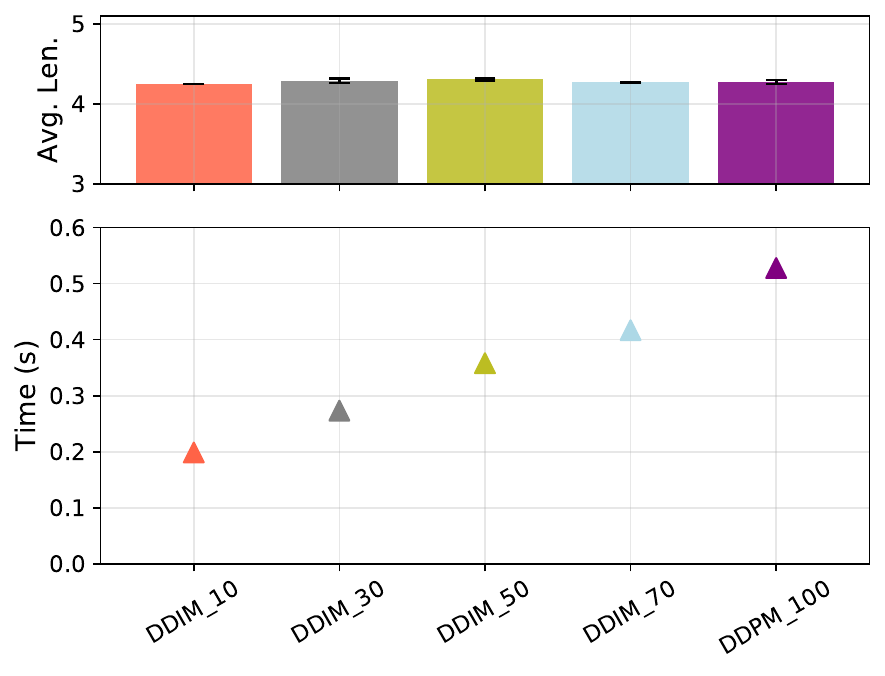}
            \caption{\textbf{Planning with the DDIM noise scheduler reduces inference time without degrading performance.} Average successful sequence length in LH-MTLC with standard error when sampling with the DDIM noise scheduler (10, 30, 50, and 70 diffusion steps) compared to DDPM (100 diffusion steps, matching training conditions). We also report the mean inference time for a single action prediction, including both planning and control, for each noise scheduler. Mean and standard error are reported over 3 seeds.}
        \label{fig:ddim_vs_ddpm}
\end{wrapfigure}

    A limitation faced by any hierarchical policy is that the decomposition of the policy into levels increases the inference overhead. This is especially true when using diffusion for \pihl; to generate a plan, the model must go through multiple stochastic denoising steps, creating a time bottleneck. Generating the entire plan (rather than online subgoals which requires more calls to \pihl) partially mitigates this, but it remains a limitation for real-world deployment. To address this issue, we investigated the use of faster noise schedulers during inference. HL was trained using the DDPM~\citep{ho2020DDPM} noise scheduler with 100 diffusion steps; we then evaluate the hierarchical policy using HL sampled with the DDIM~\citep{song2022DDIM} noise scheduler with 10, 30, 50, and 70 diffusion steps, setting $\eta=0$ for deterministic plan generation. While using fewer diffusion steps can hinder the quality of generation, it can significantly increase the prediction speed of the hierarchical policy. Figure~\ref{fig:ddim_vs_ddpm} (top) reports the mean results (Avg. Len.) over 3 seeds obtained with DDIM with different numbers of diffusion steps versus DDPM, along with the time in seconds for predicting one action (bottom) (this also includes the time for updating the physical simulator, though this is negligible). We ran these experiments on the same machine with one V100 GPU. We observe that while using DDIM with fewer diffusion steps significantly accelerates action prediction (up to $\times 2.7$ for DDIM 10), the performance does not degrade. In fact, we can assume that while using DDIM with fewer diffusion steps hinders generation at the pixel level, it does not impact the guidance provided to LL within the hierarchical policy.

\subsection{Reward function}
\label{app:reward_fun}

\method relies on a binary success signal to filter collected rollouts, which is straightforward to obtain in simulation but can become a bottleneck for real-world deployment. Importantly, however, \method requires no dense reward engineering -- only a binary success/failure label per trajectory -- which significantly lowers the barrier compared to standard RL-based fine-tuning approaches (see related work Section~\ref{sec:related_work}).

Recent advances in Vision-Language-Models~\citep{duan2024aha, grislain2026ifailsense} as binary success detectors offer a promising avenue toward fully automated real-world training. These systems take as input a visual observation of the robot and the language instruction given to the policy, and classify the trajectory as successful or failed with respect to that instruction -- precisely the signal \method requires. The simple filtering mechanism of Stage 2 (Section~\ref{subsec:training}) is compatible with such systems, making automated real-world deployment of \method a realistic extension.

\newpage

\end{document}